\documentclass[10pt,journal,compsoc]{IEEEtran}
\ifCLASSOPTIONcompsoc
  \usepackage[nocompress]{cite}
\else
  \usepackage{cite}
\fi
\ifCLASSINFOpdf
\else
\fi
\usepackage{graphicx}

\usepackage{multirow} 
\usepackage{tikz}
\usepackage{comment}
\usepackage{amsmath,amssymb} 
\usepackage{color}

\usepackage{subfig, mathtools, caption, float}
\usepackage[pagebackref=true,breaklinks=true,letterpaper=true,colorlinks,bookmarks=false]{hyperref}

\def\best{\bf \cellcolor[gray]{0.85}}
\def\secbest{\cellcolor[gray]{0.92} }

\usepackage{color}
\usepackage{colortbl}
\newcommand{\bgGray}[1]{\cellcolor[gray]{0.85} #1}

\usepackage{array}
\newcolumntype{C}[1]{>{\centering\arraybackslash}m{#1}}
\newcolumntype{R}[1]{>{\raggedleft\arraybackslash}m{#1}}
\newcolumntype{P}[1]{>{\raggedright\arraybackslash}p{#1}}
\newcolumntype{M}[1]{>{\centering\arraybackslash}m{#1}}

\newcommand{\ie}{\textit{i}.\textit{e}.}
\newcommand{\eg}{\textit{e}.\textit{g}.}

\begin{document}
%
\title{A Holistically-Guided Decoder for Deep Representation Learning with Applications to Semantic Segmentation and Object Detection}

\author{Jianbo Liu, Sijie Ren, Yuanjie Zheng, 
	   Xiaogang Wang and Hongsheng Li
\IEEEcompsocitemizethanks{\IEEEcompsocthanksitem J. Liu, X. Wang and H. Li are with the Department
of Electrical Engineering, The Chinese University of Hong Kong, Hong Kong, China.

E-mail: \{liujianbo@link, xgwang@ee, hsli@ee\}.cuhk.edu.hk

\IEEEcompsocthanksitem J. Ren is with SenseTime Research.

E-mail: rensijie@sensetime.com

\IEEEcompsocthanksitem Y. Zheng is with School of Information Science and Engineering at Shandong Normal University.

} 

}

\IEEEtitleabstractindextext{%

\begin{abstract}
Both high-level and high-resolution feature representations are of great importance in various visual understanding tasks. 
To acquire high-resolution feature maps with high-level semantic information, one common strategy is to adopt dilated convolutions in the backbone networks to extract high-resolution feature maps, such as the dilatedFCN-based methods for semantic segmentation. However, due to many convolution operations are conducted on the high-resolution feature maps, such methods have large computational complexity and memory consumption. 
To balance the performance and efficiency, there also exist encoder-decoder structures that gradually recover the spatial information by combining multi-level feature maps from a feature encoder, such as the FPN architecture for object detection and the U-Net for semantic segmentation.
Although being more efficient, the performances of existing encoder-decoder methods for semantic segmentation are far from comparable with the dilatedFCN-based methods.
In this paper, we propose one novel holistically-guided decoder which is introduced to obtain 
the high-resolution semantic-rich feature maps via the multi-scale features from the encoder.
The decoding is achieved via novel holistic codeword generation and codeword assembly operations, which 
take advantages of both the high-level and low-level features from the encoder features. 
With the proposed holistically-guided decoder, we implement the EfficientFCN architecture for semantic segmentation and HGD-FPN for object detection and instance segmentation.
The EfficientFCN achieves comparable or even better performance than state-of-the-art methods with only 1/3 of their computational costs for semantic segmentation on PASCAL Context, PASCAL VOC, ADE20K datasets. Meanwhile, the proposed HGD-FPN achieves $>2\%$ higher mean Average Precision (mAP) when integrated into several object detection frameworks with ResNet-50 encoding backbones. 

\end{abstract}

}

\maketitle

\IEEEdisplaynontitleabstractindextext

%
\IEEEpeerreviewmaketitle

\IEEEraisesectionheading{\section{Introduction}\label{sec:introduction}}

%
%
%
%
\IEEEPARstart{G}{enerating} high-resolution feature maps with rich semantic information is one of the fundamental components in modern deep neural networks with a wide variety of applications on semantic segmentation, instance segmentation, and object detection. Such tasks all require generating high-resolution and semantic-rich feature maps (e.g., 1/8 or even 1/4 of the input spatial size), which are then fed into the final segmentation or detection heads to generate the final predictions. However, given an input image, the general neural networks first perform pooling and strided convolutions to gradually decrease the spatial sizes of the feature maps and then to increase the receptive fields of the downsampled feature maps.

There are two common strategies for obtaining or recovering high-resolution feature maps. The first type of networks utilize dialated convolution 
\cite{chen2017deeplab,chen2017rethinking,yu2017dilated} to directly avoid decreasing the spatial sizes of the feature maps too much. The second type of networks adopts the encoder-decoder architectures \cite{ronneberger2015u} that gradually upsample lower-resolution feature maps via deconvolution or nearest/bilinear upsampling to generate feature maps of higher resolutions. However, both strategies have apparent limitations. 

\begin{figure*}[!t]
\centering
\begin{center}
\begin{tabular}{C{8.0cm}C{8.0cm}}
    \includegraphics[height=4.7cm]{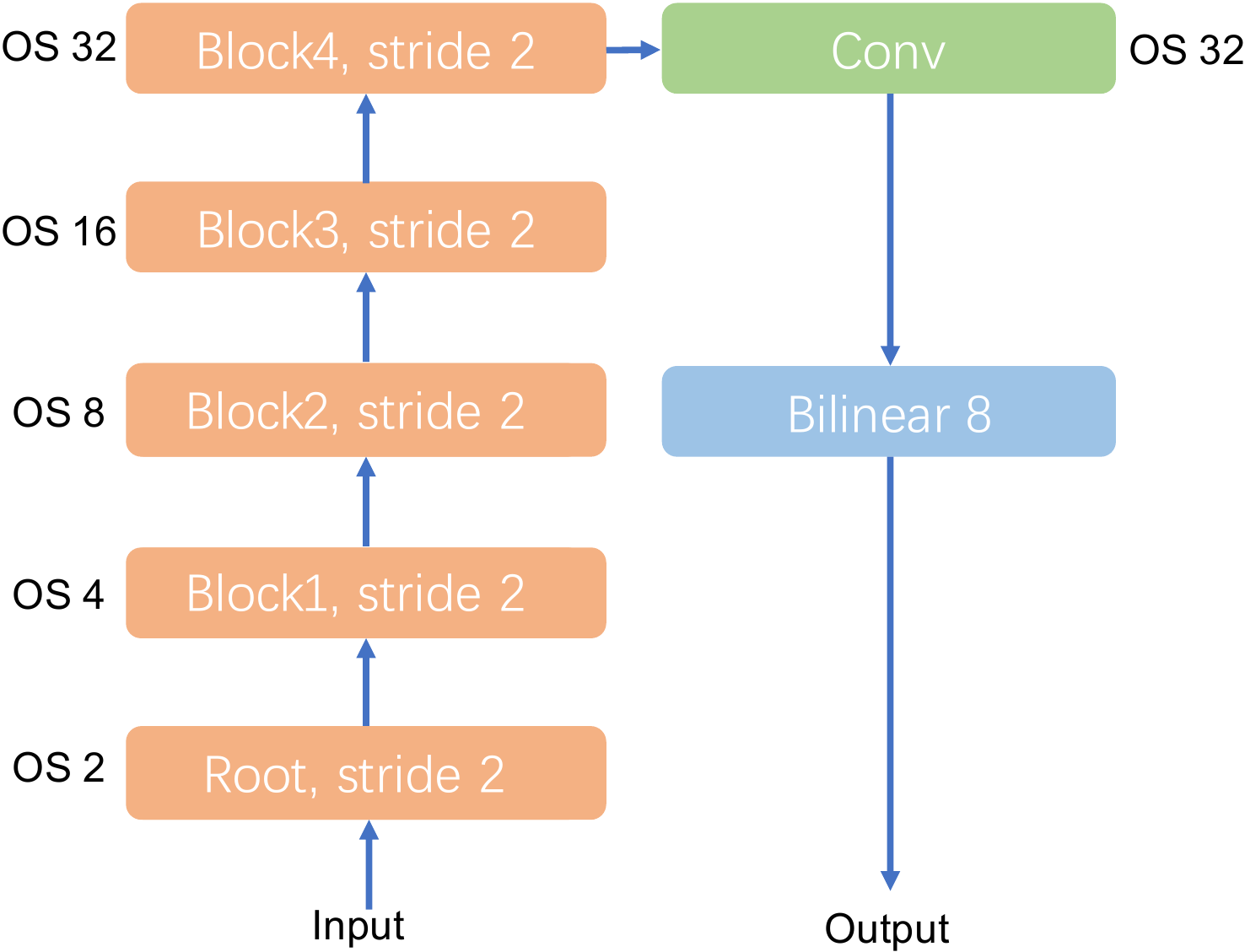} &
    \includegraphics[height=4.7cm]{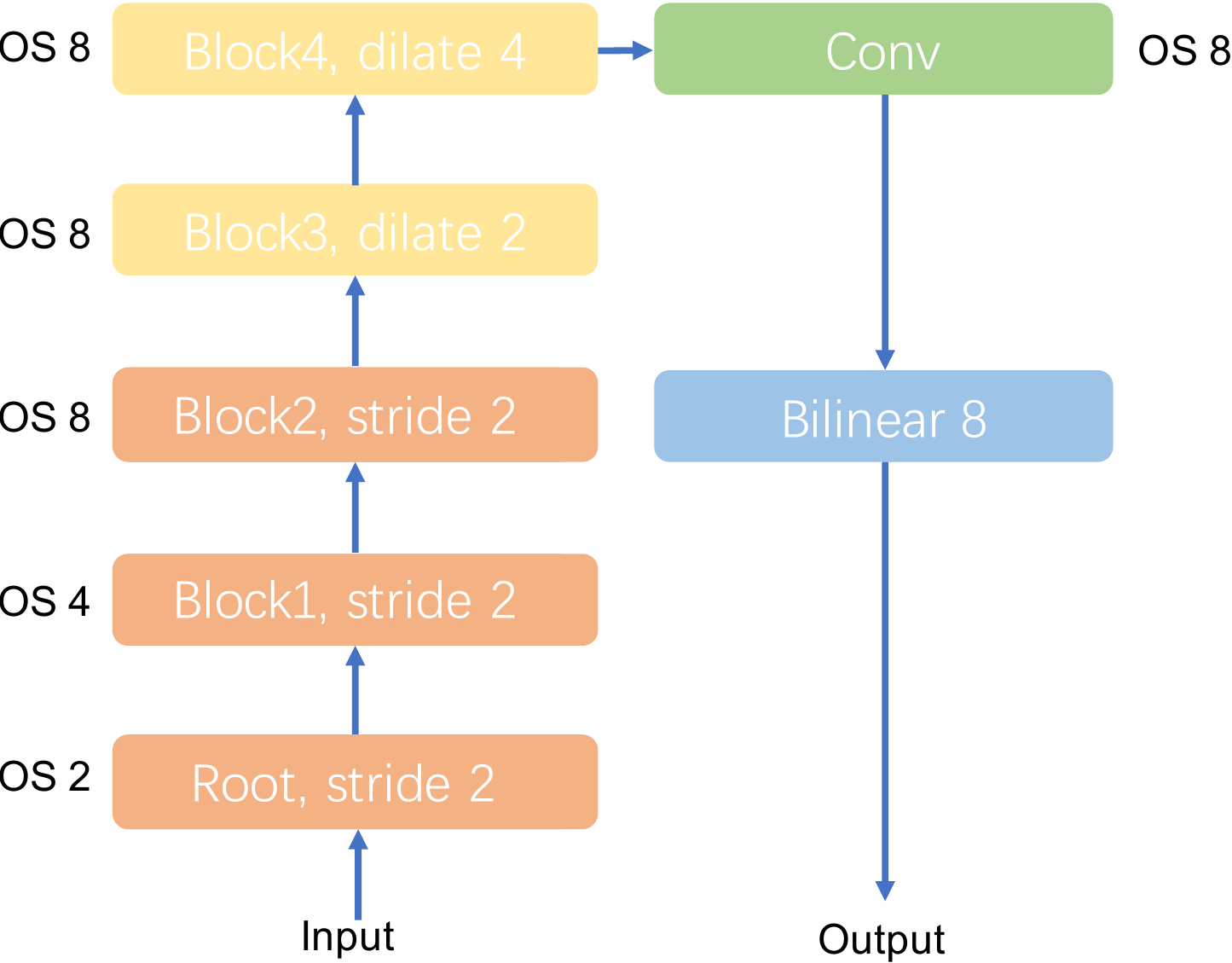} \\
    \centering (a) FCN & (b) DilatedFCN \\ \\
    \includegraphics[height=4.7cm]{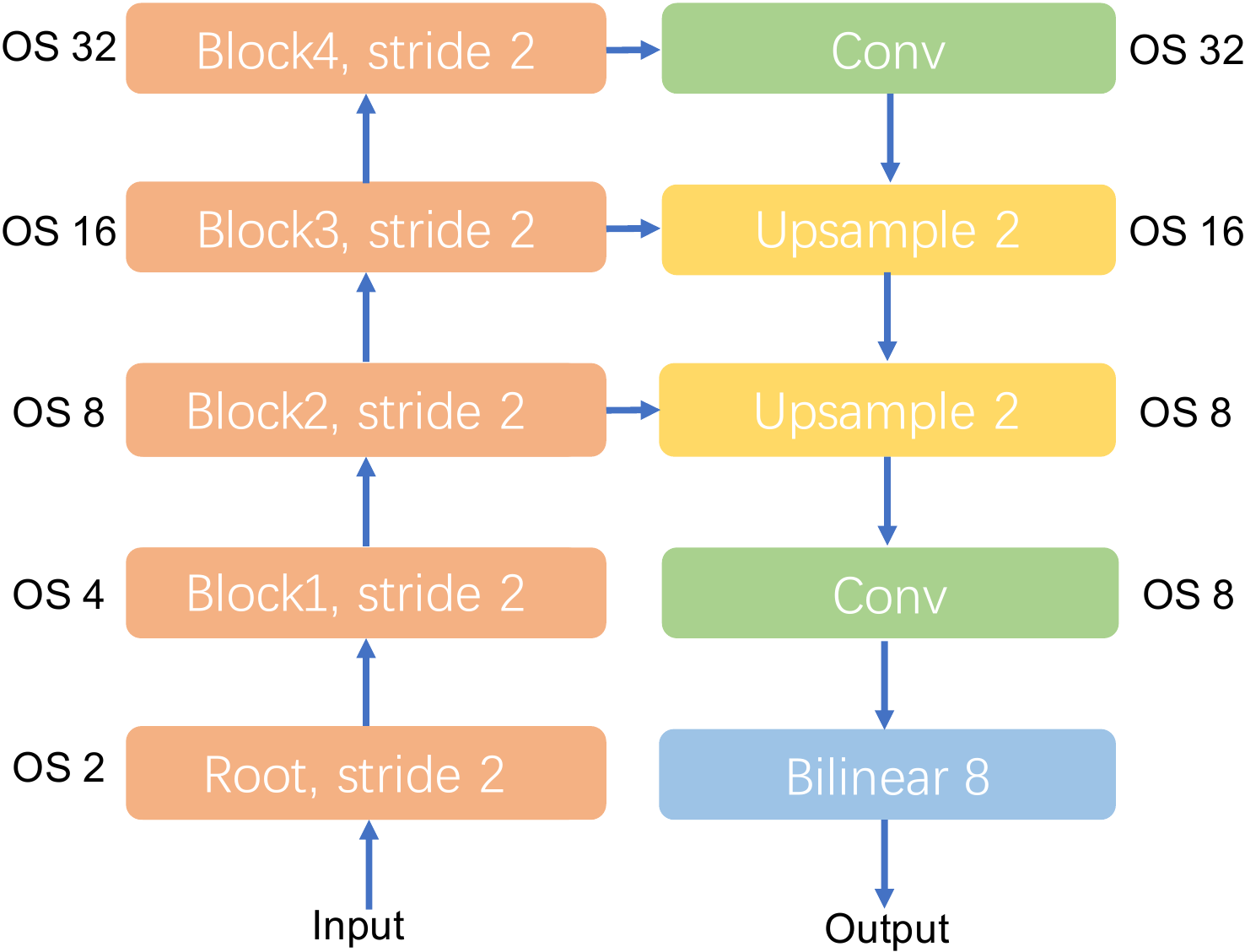} &
    \includegraphics[height=4.7cm]{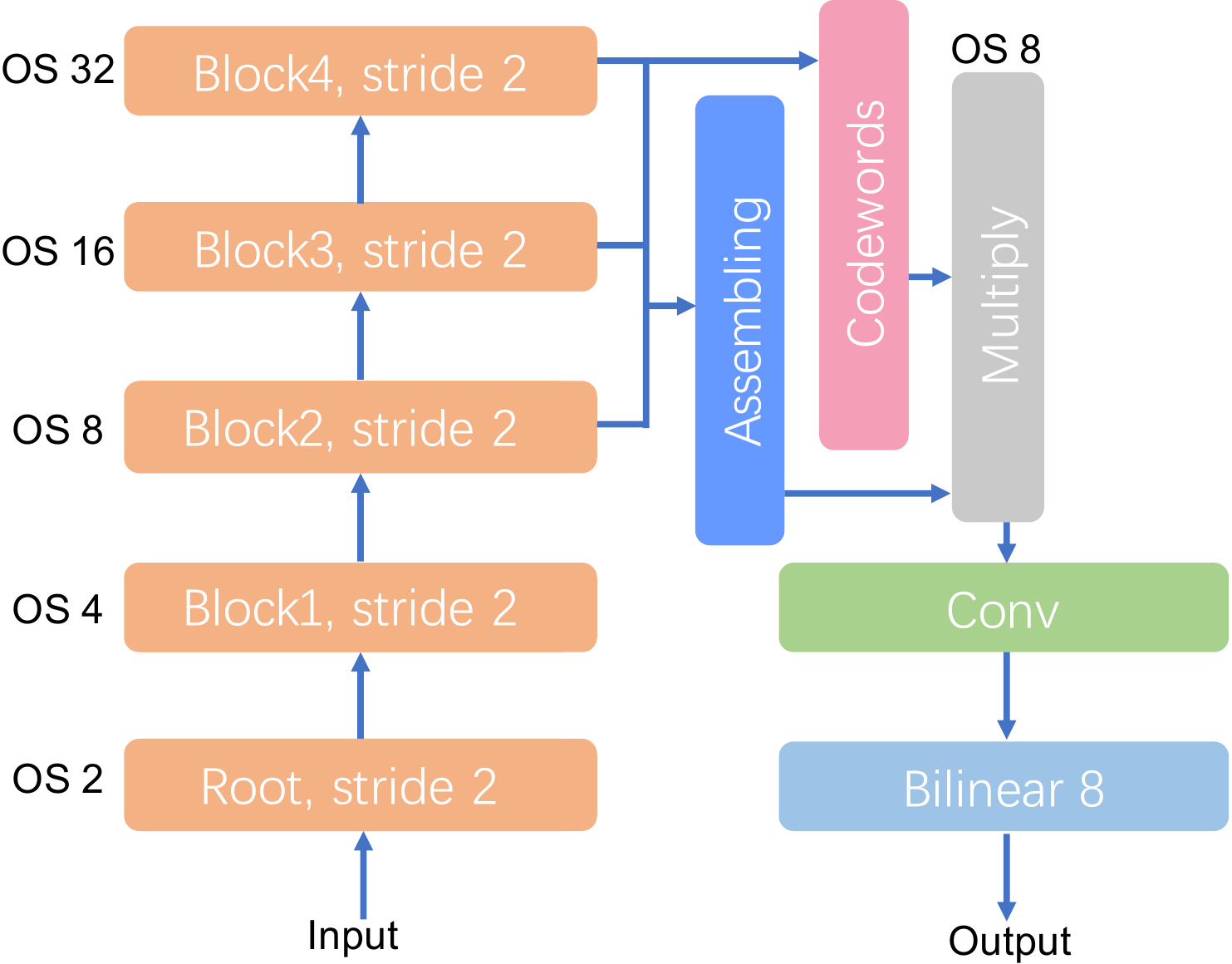} \\
    \centering (c) Encoder-Decoder & (d) EfficientFCN 
\end{tabular}
\end{center}
\caption{Different architectures for semantic segmentation. (a) the original FCN with output stride (OS)=32. (b) DilatedFCN based methods sacrifice efficiency and exploit the dilated convolution with stride 2 and 4 in the last two stages to generate high-resolution feature maps. (c) Encoder-decoder methods employ the U-Net structure to recover the high-resolution feature maps. (d) Our proposed EfficientFCN with holistic codeword generation and codeword assembly for achieving high-resolution feature upsampling in semantic segmentation.}
\label{fig:framework_img0}
\end{figure*}

The dilated convolution (or atrous convolution) is a powerful technique that can effectively enlarge the receptive field (RF) while maintaining the resolution of higher-level feature maps.
By utilizing multiple dilation rates, the dilated convolution can capture multi-scale context information without decreasing the resolution. Combining the dilated convolution with Fully Covolutional Network (FCN), the dilatedFCN based methods (shown in Fig.~\ref{fig:framework_img0}(b)) dominate state-of-the-art methods for semantic segmentation. 
As shown in Fig.~\ref{fig:framework_img0}(b), by removing the downsampling
operations and replacing convolution with the dilated convolution
in the later blocks, the dilatedFCN backbone generates the final feature maps of output stride 8 (OS=8). Based on the dilatedFCN backbone, many researcher proposed their context modeling modules to further capture the long-range context information \cite{yu2017dilated,chen2017deeplab,Zhang_2018_CVPR,he2019adaptive,Zhang_2019_CVPR}.
Despite the superior performance and no extra parameters introduced from dilated convolutions, the high-resolution feature representations require very high computational cost and memory consumption. 
For instance, for a 512$\times$512 image and a ResNet-101 backbone encoder, the computational cost of the encoder increases from 44.6 GFlops to 223.6 GFlops by adopting the dilated convolution of strides 2 and 4 into the last two blocks.

The encoder-decoder architecture, as shown in Fig.~\ref{fig:framework_img0}(c), is widely adopted in semantic segmentation and
object detection, \eg, U-Net \cite{ronneberger2015u} and SegNet \cite{SegNet}, and Feature Pyramid Network (FPN) \cite{lin2017feature}. 
The decoder networks in semantic segmentation or the FPN module in object detection gradually upsample and generate the high-resolution feature maps by aggregating multi-scale feature maps from the encoder or the backbone network. 
However, on the one hand, the fine-grained structural details are already lost in the topmost high-level encoder feature maps of output stride 32 (OS=32). Even with the skip connections, low-level high-resolution feature maps cannot provide semantic-rich features for achieving accurate segmentation or detection. On the other hand, existing feature upsampling operators mainly utilize bilinear upsampling or deconvolution to increase the resolution of the high-level feature maps. These operations can only aggregate local information for feature upsampling. The feature representation at each location of the upsampled feature maps is therefore created from a small receptive filed, which limits the contextual scope of the features.

\begin{figure*}[t]
\centering
\begin{center}
\begin{tabular}{C{3.1cm}C{3.1cm}C{4.0cm}C{4.2cm}}
    \includegraphics[height=4.9cm]{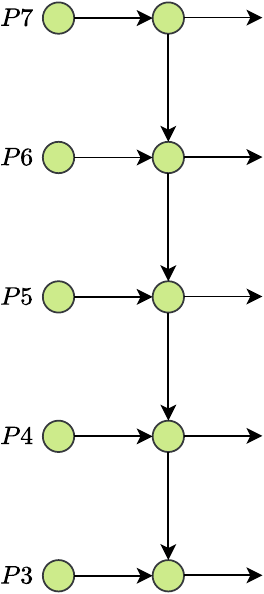} &
    \includegraphics[height=4.9cm]{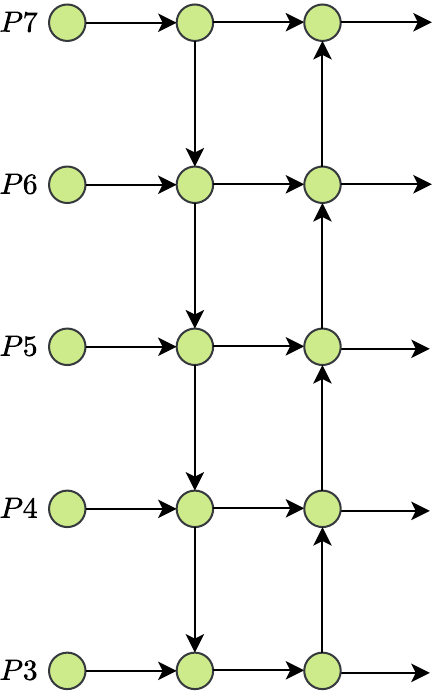} &
    \includegraphics[height=5.0cm]{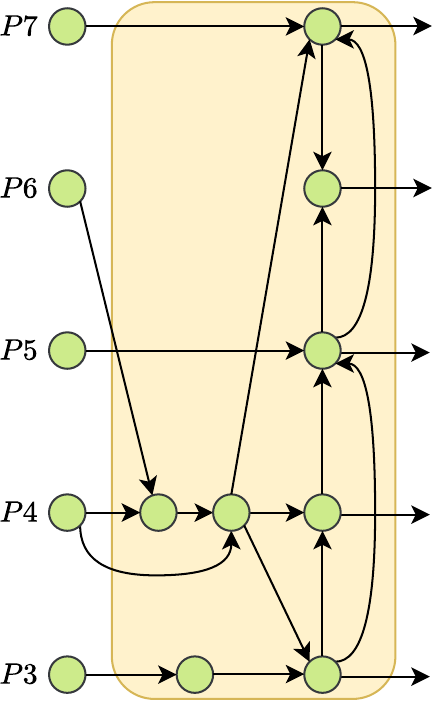} &
    \includegraphics[height=4.9cm]{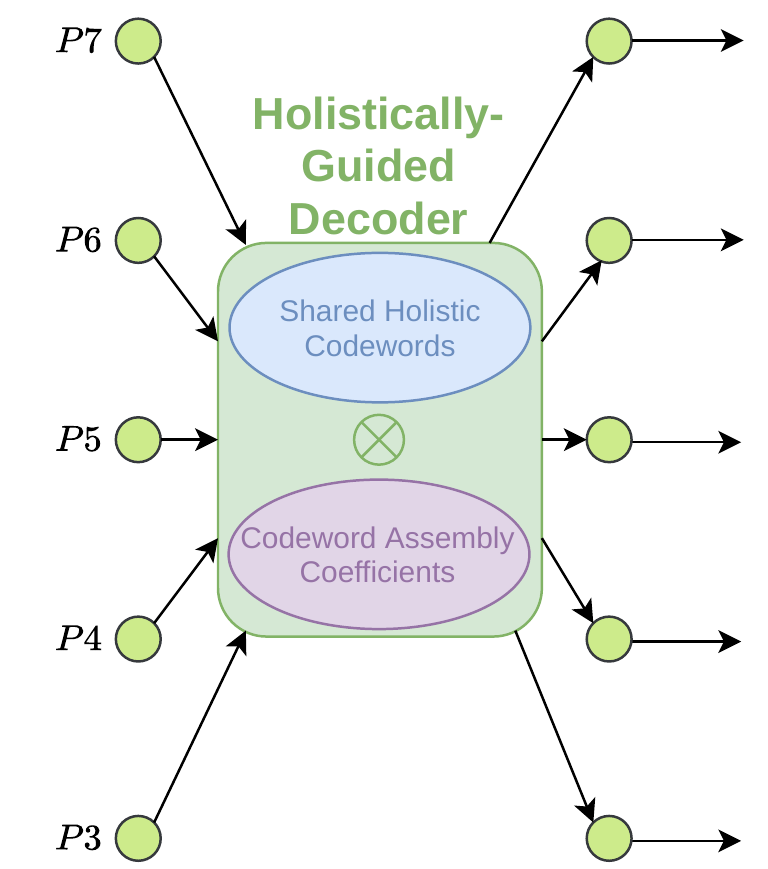} \\
    \centering (a) FPN \cite{lin2017feature} & (b) PA-FPN \cite{liu2018path}  & (c) NAS-FPN \cite{ghiasi2019fpn} & (d) our proposed HGD-FPN
\end{tabular}
\end{center}
\caption{Feature pyramid netwok (FPN) variants for object detection. (a) FPN \cite{lin2017feature} generates the feature pyramids for object detection and instance segmentation through the top-down pathway and lateral connections. (b) PA-FPN \cite{liu2018path} extends FPN by introducing an extra bottom-up pathway to enhance information fusion. 
(c) NAS-FPN \cite{ghiasi2019fpn} applies neural architecture search to automatically discover how to merge features at different scales. (d) Our HGD-FPN adopts the proposed holistically-guided decoder to modify the original FPN architecture for generating more discriminative feature pyramids.}
\label{fig:framework_fpn_variants}
\end{figure*}

To tackle the challenges in both types of networks, we propose the Holistically-guided Decoder (HGD)
(as illustrated in Fig.~\ref{fig:framework_img0}(d)) to effectively conduct feature upsampling guided by holistic context information from multi-scale feature maps.


Given the multi-level feature maps from the last three blocks from a common backbone encoder (e.g., ResNet-101), the proposed holistically-guided decoder takes the advantages of both high-level but low-resolution (OS=32) feature maps and also mid-level high-resolution feature maps (OS=8, OS=16). Intuitively, the higher-resolution but mid-level feature maps contain more fine-grained structural information, which is beneficial for spatially guiding feature upsampling; the lower-resolution but high-level feature maps contain more semantic-rich information, which are more suitable to encode the global context. Our HGD first generates a series of holistic codewords to summarize different global context of the input image from the low-resolution feature maps (OS=32). Those codewords are then properly assembled in a high-resolution grid to form the upsampled feature maps with rich semantic information. Following this principle, our HGD generates assembly coefficients from the mid-level high-resolution feature maps (OS=8, OS=16) to guide the linear assembly of the holistic codewords at each high-resolution spatial location to achieve feature upsampling.

To demonstrate the effectiveness and generalization of our proposed holistically-guided decoder, we apply it in semantic segmentation, object detection and instance segmentation. For semantic segmentation, we propose EfficientFCN by directly replacing conventional decoder in encoder-decoder architectures with the proposed HGD to generate semantically strong feature maps of OS=8. 
For object detection and instance segmentation, the proposed HGD are integrated into
the FPN module to implement the HGD-FPN module. Since the original FPN module needs to output multiple feature maps with different scales, our HGD-FPN generates these feature maps by predicting multiple assembly coefficients with different scales and using one shared holistic codewords. In such a way, the feature pyramids generated by our proposed HGD-FPN consider the holistic semantic information
from the multi-stage feature maps to boost the performance of object detection and instance segmentation.

Our primary contributions could be summarized into fourfold. (1) We propose a novel holistically-guided decoder, which can efficiently generate the high-resolution feature maps considering holistic contexts of the input image. (2) Because of the light weight and high performance of the proposed holistically-guided decoder, our proposed EfficientFCN for semantic segmentation can adopt the encoder without any dilated convolution but still achieve superior performance. (3) Our EfficientFCN achieves competitive (or better) results compared with the state-of-the-art dilatedFCN based methods on the PASCAL Context, PASCAL VOC, ADE20K datasets, with 1/3 fewer FLOPS. (4) For object detection and instance segmentation, compared to the conventional FPN, our proposed HGD-FPN achieves at least $2\%$ higher mean Average Precision (mAP) for the classical Faster-RCNN \cite{ren2015faster},  RetinaNet \cite{lin2017focal}, FreeAnchor \cite{zhang2019freeanchor} and Mask-RCNN \cite{he2017mask} frameworks with the ResNet-50 encoding backbone.

\section{Related Work}

\noindent \textbf{DilatedFCN based Methods for Semantic Segmentation} 
The Deeplab V2 \cite{chen2017deeplab,chen2017rethinking} proposed to exploit dilated convolution in
the backbone to generate high-resolution feature maps, which increases the
output stride from 32 to 8. However, the dilated convolution in the last two layers of the backbone adds huge additional computation and leaves large
memory footprint. Based on the dilated convolution backbone, many works \cite{Zhang_2019_CVPR,fu2019dual,Fu_2019_ICCV,he2019dynamic} continued to apply different strategies as the segmentation heads to acquire the context-enhanced feature maps. PSPNet \cite{zhao2017pyramid} utilized the Spatial Pyramid Pooling (SPP) module to increase the receptive field. EncNet \cite{Zhang_2018_CVPR} proposed an encoding layer to predict a feature re-weighting vector from the global context and selectively high-lights 
class-dependent feature maps. 
CFNet \cite{Zhang_2019_CVPR} exploited an aggregated co-occurrent feature (ACF) module to
aggregate the co-occurrent context by the pair-wise similarities in the feature space.
Gated-SCNN\cite{takikawa2019gated} proposed to use a new gating mechanism to connect the intermediate layers and a new loss function that exploits the duality
between the tasks of semantic segmentation and semantic
boundary prediction.
DANet \cite{fu2019dual} proposed to use two attention modules with the self-attention mechanism to aggregate features from
spatial and channel dimensions respectively.
ACNet \cite{Fu_2019_ICCV} applied a dilated ResNet as the backbone and combined the encoder-decoder strategy for the 
observation that the global context from high-level features
helps the categorization of some large semantically confused
regions, while the local context from lower-level visual features helps to generate sharp boundaries of clear details.
DMNet \cite{he2019dynamic} generated a set of dynamic filters of different
sizes from the multi-scale neighborhoods for handling the scale variations of objects for semantic segmentation.
Although these works further improve the performances on different benchmarks, these proposed heads 
still adds extra computational costs to the already burdensome encoder.


\noindent \textbf{Encoder-Decoder Architecture for Semantic Segmentation} 
Another type of methods focus on efficiently acquire the high-resolution semantic feature maps via the encoder-decoder architectures. Through
the upsampling operations and the skip connections, the encoder-decoder
architecture \cite{ronneberger2015u} can gradually recover the high-resolution feature maps for 
segmentation. DUsampling \cite{tian2019decoders} designed a data-dependent upsampling module based on fully 
connected layers for constructing the high-resolution feature maps from the low-resolution 
feature maps. FastFCN \cite{wu2019fastfcn} proposed a Joint Pyramid Upsampling (JPU) method via multiple 
dilated convolution to generate the high-resolution feature maps.
One common drawback of these methods is that the feature at each location of the upsampled high-resolution feature maps is 
constructed via only local feature fusion. Such a property limits their performance in 
semantic segmentation, where global context is important for the final performance.

\noindent \textbf{Multi-scale Feature Representations for Object Detection}
Deep learning based object detection methods have shown dramatic improvements in both
accuracy and speed recently.
Although many deep learning based object detection algorithms, such as Fast R-CNN \cite{girshick2015fast} and Faster RCNN \cite{ren2015faster} could be conducted on 
a single layer's feature maps to detect objects, detecting objects of a large variety of scales
and aspect ratios from a single-scale feature map is still challenging.
To remedy the issue, an intuitive way is to leverage multi-scale feature representations.
There are two types of strategies to generate multi-scale feature representations: multi-scale image pyramids \cite{singh2018analysis, singh2018sniper} and multi-scale
feature pyramids \cite{lin2017feature,liu2018path,pang2019libra,wang2019carafe}.  SNIP \cite{singh2018analysis} is one representative method that exploits multi-scale image pyramids, which proposed a scale normalization 
method that selectively trains the objects of appropriate sizes in each image scale. To improve the effectiveness of the 
multi-scale training, SNIPER \cite{singh2018sniper} was proposed to select context regions around the ground-truth instances and
sampling background regions for each scale during training.
However, the inference time of these methods is too long to make these image pyramid-based methods less favorable for practical applications.

Different with the image pyramids, as shown in Fig. \ref{fig:framework_fpn_variants}, feature-based pyramids aim to construct the multi-scale feature representations 
directly by exploiting the multi-layer features from the encoder network with only one input image.
As one of the pioneering works, feature pyramid network (FPN) \cite{lin2017feature} introduced a top-down pathway and lateral connections between bottom-layer and top-layer features to create multi-scale feature maps.
Motivated by FPN, PANet \cite{liu2018path} further enhanced the feature hierarchies by additional bottom-up path
augmentation and proposed adaptive feature pooling to aggregate features from all levels.
Libra RCNN \cite{pang2019libra} proposed the blanced FPN architecture which applies one non-local module to produce one balanced semantic features.
CARAFE \cite{wang2019carafe} incorporated a cascade structure and dynamic filtering to improve the quality of object proposals and detection results.
More recently, NAS-FPN \cite{ghiasi2019fpn} leveraged neural architecture search to automatically design feature network topology. Although it
achieves better performance, NAS-FPN \cite{ghiasi2019fpn} requires thousands of GPU hours during search, and the resulting feature network is irregular and thus difficult to interpret.
Feature pyramids grids (FPG) \cite{chen2020feature} proposed a deep multi-pathway feature pyramid network by representing the feature scale-space as a regular grid of parallel pyramid
pathways. However, similar to encoder-decoder methods for semantic segmentation, most of these
FPN-variants reconstruct the multi-scale feature representations in a local feature fusion manner and
cannot efficiently fuse the global semantic information into the final multi-stage feature maps.
In this paper, our proposed HGD can be integrated with the FPN module to implement our HGD-FPN, which is able to generate discriminative multi-scale feature representations that contain the holistic semantic information from the multi-level feature maps.

\begin{figure*}[tb]
    \centering
    \subfloat{\includegraphics[width=18.0cm]{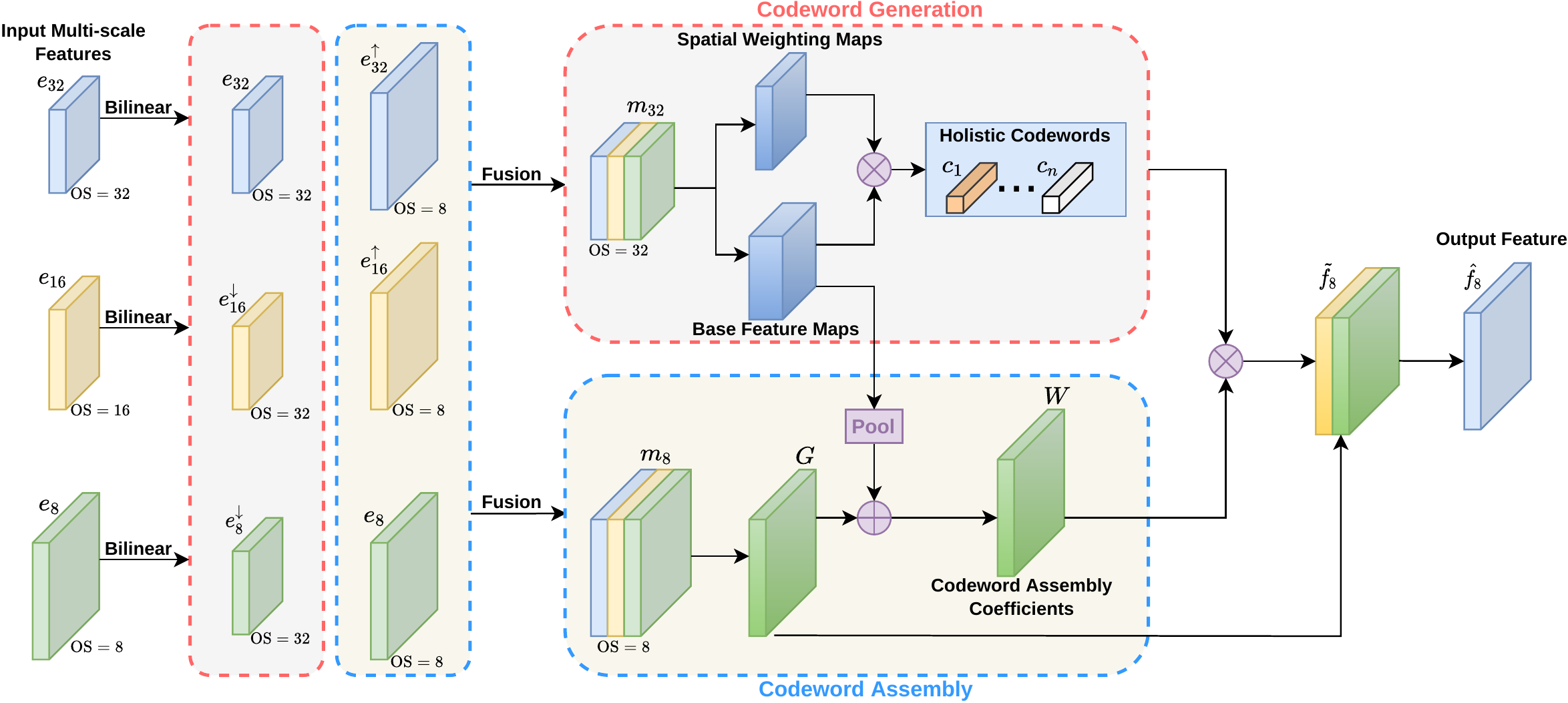}}
\caption{Illustration of the proposed holistically-guided decoder architecture, which is used in our EfficientFCN method for semantic segmentation. It consists of three main components. Multi-scale feature fusion fuses multi-scale features to obtain OS=8 and OS=32 multi-scale feature maps. Holistic codeword generation results in a series of holistic codewords summarizing different aspects of the global context. High-resolution feature upsampling can be achieved by codeword assembly. For semantic segmentation, the symbols $\uparrow$ and $\downarrow$ denote the bilinear upsampling and downsampling
operations, respectively.}
\label{fig:framework_img1}
\end{figure*}


\section{Methods}

\subsection{Overview}

In state-of-the-art Deep Convolutional Neural Network (DCNN) backbones, the high-resolution mid-level feature maps
at earlier stages can better encode fine-grained structures, while the
low-resolution high-level features at later stages are generally more
discriminative for category prediction. 
They are essential and complementary information sources for generating high-resolution feature representations, which are required for 
achieving accurate semantic segmentation and object detection.
To combine the strengths of both high-level and low-level features, encoder-decoder structures and the Dilated Fully Convolutional Network (DilatedFCN)-based 
\cite{yu2017dilated,chen2017deeplab,Zhang_2018_CVPR,he2019adaptive,Zhang_2019_CVPR,ronneberger2015u} methods
were developed to generate high-resolution feature maps with semantic-rich
information from the multi-scale feature maps. Take an encoder-decoder structure as an example.
To generate the final feature map $\tilde{f}_8$ of OS=8 for mask prediction, a three-stage encoder-decoder first upsamples the topmost encoder feature map $f_{32}$ of OS=32 to generate OS=16 feature maps $f_{16}$. The OS=16 feature maps $e_{16}$ of the same size from the encoder is either directly concatenated as $[f_{16}; e_{16}]$ (U-Net \cite{ronneberger2015u}) or summed as $f_{16}+e_{16}$ followed by some $1\times 1$ convolutions to generate the upsampled OS=16 feature maps, $\tilde{f}_{16}$. The same upsampling + skip connection procedure repeats again for $\tilde{f}_{16}$ to generate $\tilde{f}_{8}$. The upsampled features $\tilde{f}_8$ eventually contain both mid-level and high-level information to some extent and can be used to generate the segmentation masks.

However, since bilinear upsampling and deconvolution layer in classical decoders are local operations with limited receptive fields. We argue that they are incapable of exploiting important global context of the input image, which are crucial for achieving accurate segmentation. Although there were existing attempts \cite{Zhang_2018_CVPR,hu2018squeeze} on using
global context to re-weight the contributions of different channels of the
feature maps either in the backbone \cite{Zhang_2018_CVPR} or in the upsampled feature
maps \cite{hu2018squeeze}. This strategy only scales each feature channel but
maintains the original spatial size and structures. Therefore, it is incapable
of generating high-resolution semantic-rich feature maps to identify fine-grained image structures.

To solve this drawback, we propose a novel Holistically-guided Decoder (HGD), as shown in Fig. \ref{fig:framework_img1},
which decomposes the feature upsampling process into the generation of a series of holistic codewords from multi-level feature maps to capture global contexts, and linearly assembling codewords at each spatial location for achieving semantic-rich feature upsampling. Such a decoder can exploit the global contextual information to effectively guide the feature upsampling process and is able to recover
fine-grained details. 
Thus, our proposed HGD can be integrated into the deep learning frameworks, where the high-resolution
feature representations are needed, such as the semantic segmentation, instance segmentation,
and object detection.
In the following subsections, we first introduce our proposed HGD generating
the high-resolution (OS=8) semantic-rich feature maps from multi-scale feature maps of OS=8, 
OS=16 and OS=32, which can be directly used as the decoder module into the encoder-decoder 
architectures for semantic segmentation. As object detection and instance segmentation are conducted on feature pyramids,
we show how to generalize our HGD module with some minor modifications to implement one novel feature pyramids generation method HGD-FPN, which 
can encode discriminative multi-scale feature representations to significantly improve the performance of popular
detection and instance segmentation frameworks, including Faster R-CNN \cite{ren2015faster}, Mask-RCNN \cite{he2017mask}, RetinaNet \cite{lin2017focal} and FreeAnchor \cite{zhang2019freeanchor}.

\subsection{Holistically-guided Decoder}
\label{sec:methods_hgd}

\noindent \textbf{Multi-scale features fusion.}
Given the multi-scale feature maps from the encoder, we aim at generating a series of holistic codewords to encode different aspects of the input image's global context, and linear assembly coefficients for combining the holistic codewords for creating the high-resolution feature maps. Without loss of generality, we assume that the encoder generates three feature maps with different scales (OS=8, 16 and 32) and the decoder generates semantic-rich feature maps of scale OS=8.

Although we can directly
generate the holistic codewords from the topmost high-level OS=32 feature maps and also
directly produce the codeword assembly coefficients from the mid-level
OS=16 and OS=8 feature maps, we observe the simple fusion of multi-scale encoder feature maps can lead to better performance. 
Specifically, for the OS=8, OS=16, OS=32 encoder feature maps, we first adopt
separate $1\times 1$ convolutions to compress each of their channels to 512
for reducing the follow-up computational complexity, obtaining $e_{8} \in \mathbb{R}^{512 \times (H/8)\times (W/8)}$, $e_{16}\in \mathbb{R}^{512 \times (H/16)\times (W/16)}$,
$e_{32} \in \mathbb{R}^{512 \times (H/32)\times (W/32)}$, respectively, where $H$ and $W$ are the input image's height and width. The multi-scale fused OS=32 feature maps $m_{32}$ are
then obtained by downsampling $e_{8}$ and $e_{16}$ to the size of OS=32 and
concatenating them along the channel dimension with $e_{32}$ as $m_{32} = [
    e_{8}^\downarrow;  e_{16}^\downarrow; e_{32}] \in \mathbb{R}^{1536\times
(H/32) \times (W/32)}$, where $^\downarrow$ represents bilinear downsampling, and
$[\cdot;\cdot]$ denotes concatenation along the channel dimension. We can also obtain
the multi-scale fused OS=8 feature maps, $m_8 = [ e_{8}; e_{16}^\uparrow;
e_{32}^\uparrow] \in \mathbb{R}^{1536\times (H/8) \times (W/8)}$, in a similar
manner.

\noindent \textbf{Holistic codeword generation.}
Although the multi-scale fused feature maps $m_{32}$ are created to integrate
both high-level and mid-level features, their small resolutions make them lose
many structural details of the scene. On the other hand, because $e_{32}$ is
encoded from the deepest layer, $m_{32}$ is able to encode rich categorical
representations of the image. We therefore propose to generate a series of
unordered holistic codewords from $m_{32}$ to implicitly model different
aspects of the global context.

To generate $n$ holistic codewords, a codeword bases map $B \in
\mathbb{R}^{1024 \times (H/32)\times}$ $^{(W/32)}$ and $n$ spatial weighting
maps $A \in \mathbb{R}^{n \times (H/32)\times (W/32)}$ are first computed from
the fused multi-scale feature maps $m_{32}$ by two separate $1\times 1$
convolutions. 
For the bases map $B$, we denote $B(x,y) \in \mathbb{R}^{1024}$ as the 1024-d
feature vector at location $(x,y)$; for the spatial weighting maps $A$, we use
$A_i \in \mathbb{R}^{(H/32) \times (W/32)}$ to denote the $i$th weighting map.
To ensure the weighting maps $A$ are properly normalized, the softmax function is adopted to 
normalize all spatial locations of each channel $i$ (the $i$-th spatial feature map) as
\begin{align}
            \tilde{A}_i(x,y) = \frac{\exp(A_i(x,y))}{\sum_{\textrm{all }p,q}
            \exp(A_i(p,q))}.
\end{align}
The $i$-th codeword $c_i\in \mathbb{R}^{1024}$ can be obtained as the weighted
average of all codeword bases $B(x,y)$, \ie,
\begin{align}
            c_i = \sum_{p,q} \tilde{A}_i(p,q) B(p,q).
            \label{eq:codeword_generation}
\end{align}
In other words, each spatial weighting map $\tilde{A}_i$ learns to linearly
combine all codeword bases $B(x,y)$ from all spatial locations to form a
single codeword, which captures certain aspect of the global context. The $n$
weighting maps eventually lead to $n$ holistic codewords $C = [c_1, \dots,
c_n] \in \mathbb{R}^{1024 \times n}$ to encode high-level global features.

\noindent \textbf{Codeword assembly for high-resolution feature upsampling.}
The holistic codewords can capture various aspects of the global context of the input image.
They are perfect ingredients for generating the high-resolution
semantic-rich feature maps as they are encoded from the high-level features
$m_{32}$. However, since their structural information have  been mostly
removed during the downsampling process, we turn to use the OS=8 multi-scale fused
features $m_8$ to predict the linear assembly coefficients of the $n$
codewords at each spatial location for creating a high-resolution feature map.

\begin{figure*}[!t]
    \centering
    \subfloat{\includegraphics[width=17.0cm]{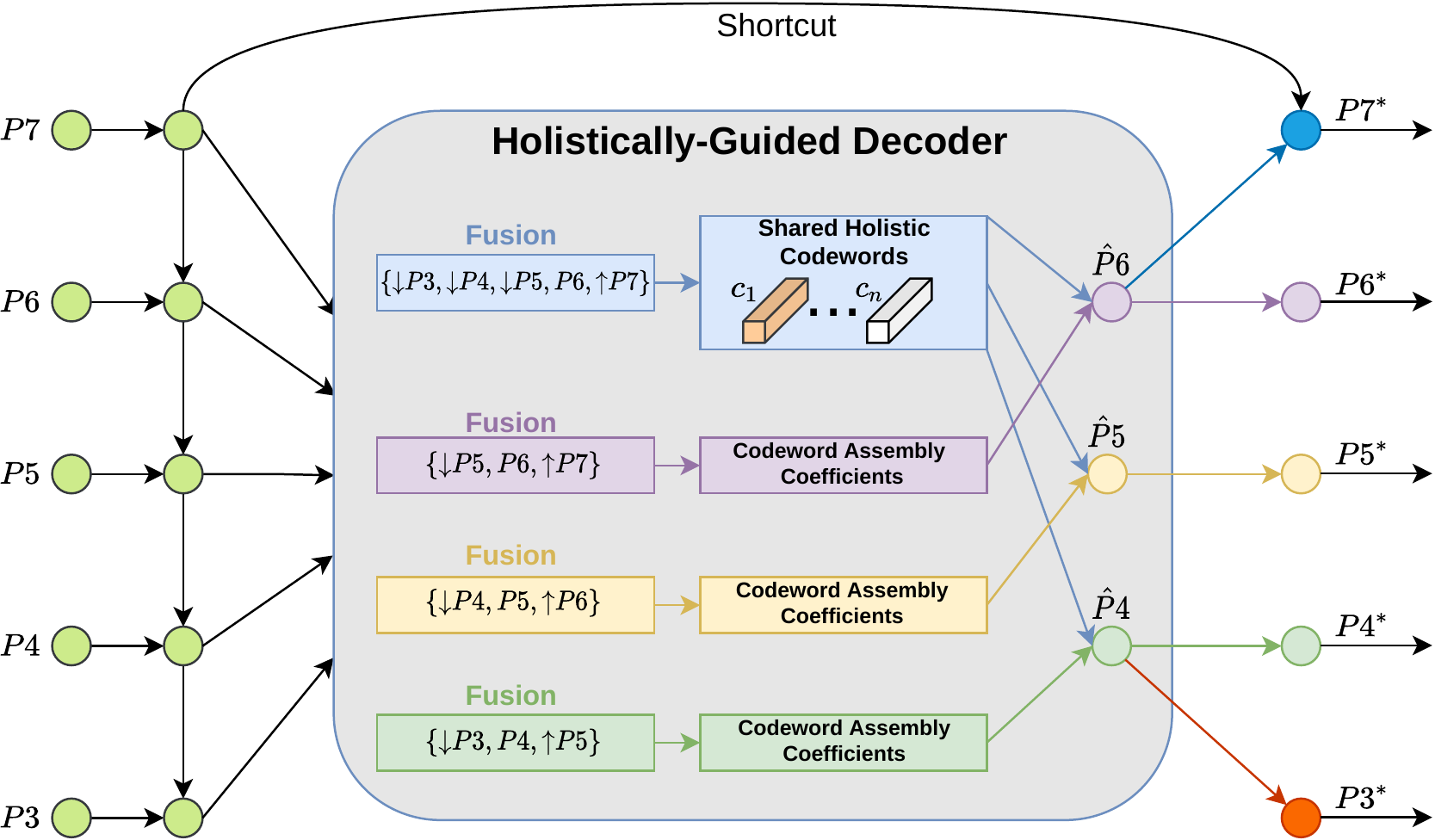}}
\caption{Illustration of the proposed HGD-FPN architecture for object detection and instance segmentation. HGD-FPN integrates our proposed holistically-guided decoder into the FPN architecture to generate
the feature pyramids. It also consists of three main components. Given the multi-scale feature maps {$P3-P7$}, the multi-scale feature fusion component fuses multi-scale features for the codeword generation and codeword assembly components respectively. 
The HGD-FPN applies the codeword generation component to produce one shared set of holistic codewords from the multi-scale feature maps {$P3-P7$}. 
Then, it exploits three codeword assembly components to generate three feature maps with different scales. Finally, the residual connections are used to generate the output feature pyramids. In our HGD-FPN for object detection, the symbols $\uparrow$ and $\downarrow$ denote the nearest upsampling interpolation and max-pooling operations, respectively.}
\label{fig:framework_hgd_fpn}
\end{figure*}

More specifically, we first create a raw codeword assembly guidance feature
map $G \in \mathbb{R}^{1024\times (H/8)\times (W/8)}$ to predict the assembly
coefficients at each spatial location, which are obtained by applying a
$1\times 1$ convolution on the multi-scale fused features $m_8$. However, the
OS=8 fused features $m_8$ have no information on the holistic codewords as they
are all generated from $m_{32}$. We therefore consider the general codeword
information as the global average vector of the codeword based map $\bar{B} \in
\mathbb{R}^{1024}$ and location-wisely add it to the raw assembly guidance
feature map to obtain the novel guidance feature map $\bar{G} = G \oplus \bar{B}$, where $\oplus$ represents
location-wise addition. Another $1\times 1$ convolution applied on the
guidance feature map $\bar{G}$ generates the linear assembly coefficients or weighting map of the $n$
codewords $W \in \mathbb{R}^{n \times (H/8) \times (W/8)}$ for all $(H/8)
\times (W/8)$ spatial locations.  By reshaping the weighting map $W$ as an $n
\times (HW/8^2)$ matrix, the holistically-guided upsampled feature
$\tilde{f}_8$ of OS=8 can be easily obtained as
\begin{align}
            \tilde{f}_8 = W^\top C.
            \label{eq:upsample}
\end{align}

Given the holistically-guided upsampled feature map $\tilde{f}_8$, we obtain the final 
upsampled feature map  $\hat{f}_8$ by concatenating the feature map $\tilde{f}_8$ with the
guidance feature map $G$, \ie~$\hat{f}_8 = [\tilde{f}_8; G]$.
Such an upsampled feature map $\hat{f}_8$ takes advantages of both $m_8$
and $m_{32}$, and contains semantic-rich and also structure-preserved features
for achieving one powerful high-resolution representation. If other output scales are needed (e.g., $\hat{f}_{16}$ of OS=16), they can be generated in the similar manner as $\hat{f}_8$.


\subsection{EfficientFCN with HGD for Semantic Segmentation}
Semantic segmentation aims to assign one categorical label to each pixel on the given image.
For predicting the semantic segmentation result, one $1\times 1$ convolution layer is adopted as the final classifier appended to the final feature maps.
Most of the state-of-the-art methods \cite{yu2017dilated,chen2017deeplab,Zhang_2018_CVPR,he2019adaptive,Zhang_2019_CVPR} use the feature maps with OS=8 to make the categorical prediction, whose results are then upsampled to the original input resolution with bilinear interpolation. 
How to obtain the high-resolution OS=8 feature maps with rich semantic information is therefore essential for achieving high segmentation accuracy.
To compared with state-of-the-art methods, in this work, we focus on utilizing the proposed HGD to generate OS=8 feature maps, although we can use it to generate feature maps of even higher resolutions.

As illustrated in Fig.~\ref{fig:framework_img0}, there exist two types of architectures to acquire the 
high-resolution OS=8 feature maps for semantic segmentation. The DilatedFCN based methods \cite{yu2017dilated,chen2017deeplab,Zhang_2018_CVPR,he2019adaptive,Zhang_2019_CVPR}
 current dominate the benchmark leaderboards and replace the traditional convolution with the dilated convolution at the last two
stages of the encoder network. However, these methods introduce very heavy computational overhead. For the traditional encoder-decoder methods \cite{ronneberger2015u, tian2019decoders}, 
the upsampling operations, bilinear interpolation or deconvolution, generate the upsampled feature maps by fusing information of only local neighborhoods, which are sub-optimal for the semantic segmentation that needs contextual understanding. 

To remedy these shortcomings, we propose our EfficientFCN architecture that integrates our HGD into the encoder-decoder architecture with commonly used encoder backbones (e.g., ResNet \cite{he2016deep}, etc.) but adopts HGD to replace existing upsampling operations in the conventional decoder.
Specifically, the encoder takes an image of interest as input for feature encoding and outputs the initial feature maps, $f_{32}, f_{16}, f_{8}$, of three scales, which are then fed into our proposed HGD to generate the OS=8 feature map $\tilde{f}_8$ as described in Section \ref{sec:methods_hgd}. 
A final $1\times 1$ convolution is then applied on the upsampled feature map $\tilde{f}_8$ to predict pixel-wise categories. 
As shown by our experiments, compared with state-of-the-art DialatedFCN-based methods that adopt dilated convolutions, our proposed EfficientFCN use only 1/3 of their FLOPS with same backbone architectures and achieve comparable or higher segmentation accuracy.

\subsection{HGD-FPN for Object Detection and Instance Segmentation}
The Feature Pyramid Network (FPN) \cite{lin2017feature} and its variants \cite{liu2018path, wang2019carafe, ghiasi2019fpn, chen2020feature} are commonly adopted to generate
multi-scale feature representations for capturing objects of different scales in object detection and instance segmentation. The FPN has been shown as one of the most important components in modern deep learning based object detectors.
Similar to the problems of encoder-decoder architectures for semantic segmentation, 
most existing FPN methods for object detection mainly generate the multi-scale feature representations from information of local neighborhoods via nearest interpolation and the (de-)convolution operator with a small and limited receptive field. Obviously, it cannot efficiently fuse global semantic information into the final multi-scale feature maps.

To better incorporate global context into the generation of multi-scale feature maps, we integrate our proposed HGD module into the classical FPN architecture and implement one novel feature pyramids method, which is named HGD-FPN, for object detection and instance segmentation. 
Unlike HGD in EfficientFCN, which only needs to generate high-resolution feature maps of a single scale OS=8, HGD-FPN needs to generate feature maps of multiple scales. Therefore, as shown in 
Fig. \ref{fig:framework_hgd_fpn}, our designed HGD-FPN architecture makes modifications to the three above mentioned operations: multi-scale encoder feature fusion, holistic codeword generation 
and codeword assembly for feature upsampling.

Specifically, for codeword generation we first fuse all multi-scale encoder feature maps $(P3 - P7)$ so that the learned holistic codewords contain semantic information of different scales. We upsample the feature map $P7$ and downsample the feature maps $\{ P3-P5\}$ to make them have the same resolution as $P6$ via nearest interpolation or max-pooling operation. The multiple feature maps are linearly combined to generate the multi-scale fused feature maps $m_{\rm code}$,
\begin{align}
m_{\rm code} = a_0  \cdot {\uparrow}{P7} + a_1  \cdot {P6} + a_2  \cdot {\downarrow}{P5} + a_3  \cdot {\downarrow}{P4} + a_4  \cdot {\downarrow}{P3},
\end{align}
where the symbols $\uparrow$ and $\downarrow$ denote the nearest upsampling interpolation and max-pooling operations, respectively. The fusion coefficients $a_i$ are learned non-negative parameters via back-propagation. To make them satisfy $\sum{a_i} = 5$, we make each coefficient go through a ReLU function as $a_i = {\rm ReLU}(a_i)$ (similar to the weighting of feature maps in \cite{tan2020efficientdet}), so that the negative coefficients would be set to zeros. After the multi-scale fused feature maps $m_{\rm code}$ are generated, we adopt the holistic codeword generation process (Eq. \eqref{eq:codeword_generation}) to generate one set of shared holistic codewords $C$ from $m_{\rm code}$. For object detection and instance segmentation, the shared holistic codewords $C$ contains 128 holistic codewords of dimension $512$.

For multi-scale feature fusion for generating codeword assembly coefficients, as different scales of the encoder feature pyramids contain semantic information of different scales, we observe that using the current scale with its two neighboring scales provide a good trade-off between computational cost and accuracy. The fused encoder feature maps corresponding to encoder feature maps $(P4-P6)$ are created as
\begin{alignat}{3}
m_4 &= r_0 \cdot  {\uparrow}{P5} &&+ r_1 \cdot {P4} &&+ r_2 \cdot  {\downarrow}{P3}, \\
m_5 &= s_0 \cdot{\uparrow}{P6} &&+ s_1  \cdot {P5} &&+ s_2  \cdot {\downarrow}{P4}, \\
m_6 &= t_0  \cdot {\uparrow}{P7} &&+ t_1 \cdot {P6} &&+ t_2  \cdot {\downarrow}{P5},
\end{alignat}
where the fusion coefficients $r_i, s_i, t_i$ are learned non-negative parameters and satisfy $\sum{r_i} = 3$, $\sum{s_i} = 3$ and $\sum{t_i} = 3$. 
$m_4, m_5, m_6$ can be used to generate the codeword assembly coefficients or weighting maps of these three scales with different and independent operations (\ie~convolution kernels are not shared across scales) following the descriptions in Section \ref{sec:methods_hgd}.
The upsampled feature maps $\hat{P}4$, $\hat{P}5$ and $\hat{P}6$ (corresponding to the scales of $P4$, $P5$, $P6$) can then be obtained by linearly combining the codewords $C$ with the linear assembly coefficients following Eq. \eqref{eq:upsample}. On the other hand, for feature maps $P3$, since they have too large resolution, to avoid too much computation, we directly upsample $\hat{P}4$ to generate $\hat{P}3$, \ie,~$\hat{P}3 = \uparrow\hat{P}4$. Similarly, as $P7$ has too small resolution, we directly downsample $\hat{P}6$ to generate $\hat{P}7$, \ie,~$\hat{P}7 = \downarrow \hat{P}6$. Residual connections are further added between $P3$-$P7$ and $\hat{P}3$-$\hat{P}7$ to generate the final output feature maps $P3^*$-$P7^*$ for the proposed HGD-FPN,
\begin{align}
	P3^* &= P3 + \hat{P}3,~~P4^* = P4 + \hat{P}4,~~P5^* = P5 + \hat{P}5, \notag \\ 
	P6^* &= P6 + \hat{P}6,~~P7^* = P7 + \hat{P}7.
\end{align}

The resulting multi-scale feature maps $P3^*$-$P7^*$ can be used similarly to those from conventional FPN for ROI pooling in existing object detection frameworks. Such multi-scale features incorporate global contexts and can help improve the final detection performance.
Following the latest work on recurrent FPN architectures \cite{chen2020feature, tan2020efficientdet, ghiasi2019fpn}, in our HGD-FPN, we find our HGD-FPN module can also be stacked for multiple $k$ times to further enhance the representation ability of the resulting multi-scale feature maps. We set $k=4$ and other different $k$'s are also explored in experiments.

We integrate the proposed HGD-FPN into three popular object detection frameworks (Faster RCNN\cite{ren2015faster}, RetinaNet \cite{lin2017focal} and FreeAnchor \cite{zhang2019freeanchor}) and one instance segmentation network (Mask RCNN \cite{he2017mask}).
ResNet is adopted as the encoder network to extract the encoder multi-scale semantic features $(P3 - P7)$ of channel dimension 256, which are input into our HGD-FPN to generate the output multi-scale feature maps.
For Faster RCNN and Mask RCNN, the feature maps $(P3 - P7)$ correspond to the feature maps of scales OS=4 to OS=64.
For RetinaNet and FreeAnchor, the feature maps $(P3 - P7)$ correspond to the feature maps of scales OS=8 to OS=128.




\section{Experiments}

In this section, we introduce the implementation details, training strategies and evaluation metrics of our experiments for three 
applications: semantic segmentation, object detection and instance segmentation.
For semantic segmentation, to evaluate our proposed EfficientFCN model, we conduct comprehensive experiments
on three public datasets PASCAL Context \cite{mottaghi2014role}, PASCAL VOC 2012 \cite{everingham2010pascal}
and ADE20K \cite{zhou2017scene}. To further evaluate the contributions of individual components in our model, we conduct detailed ablation studies on the PASCAL Context dataset. 
Then, we perform experiments on the MS COCO 2017 dataset \cite{lin2014microsoft} to demonstrate the capability of our proposed HGD-FPN module. 
In principle, our proposed HGD-FPN module can be flexibly integrated into any methods for object detection and instance segmentation by replacing the 
FPN module with our HGD-FPN. In this paper, we choose four representative frameworks, Faster RCNN\cite{ren2015faster}, RetinaNet\cite{lin2017focal}, FreeAnchor\cite{zhang2019freeanchor} and Mask RCNN\cite{he2017mask}, to verify its effectiveness.

\begin{table*}[tb]
\centering
%
\centering
\caption{Comparisons with classical encoder-decoder methods.}
\label{table:ablation_encoder_decoder}
\begin{minipage}[t]{0.6\textwidth}
\setlength{\tabcolsep}{0.3mm}{
\begin{tabular}{llllll}
\hline
\textbf{Method} & \textbf{Backbone} & \textbf{OS} &\textbf{mIoU\%} &
\textbf{Parameters (MB)} &\textbf{GFlops (G)}\\\hline\hline
FCN-32s  &  ResNet101 & 32 & 43.3 & 54.0 & 44.6 \\ 
dilatedFCN-8s  & dilated-ResNet101 & 8 & 47.2 &54.0 & 223.6 \\ 
UNet-Bilinear  & ResNet101 & 8 & 49.3 & 60.7 & 87.9 \\ 
UNet-Deconv  & ResNet101 & 8  & 49.1 & 62.8 & 93.2\\ 
\hline
EfficientFCN & ResNet101 & 8  & 55.3 & 55.8 & 69.6\\ 
\hline
\end{tabular}}
\end{minipage}
\end{table*}

\subsection{Semantic Segmentation}
\subsubsection{Implementation Details}
\noindent
\textbf{Network structure.}\
Different with the dilatedFCN based methods, which remove the stride of the last two blocks of
the backbone networks and adopt the dilated convolution with the dilation rates $2$ and $4$,
we use the original ResNet \cite{he2016deep} as our encoder backbone network. 
Thus the size of the output feature maps from the last ResBlock is $32\times$ smaller than that of
the input image. After feeding the encoder feature maps into our proposed holistic-guided decoder,
the classification is performed on the output upsampled feature map $\hat{f}_8$.
The ImageNet \cite{russakovsky2015imagenet} pre-trained weights are utilized to initialize the encoder network.
%
%
%

\noindent
\textbf{Training setting.}\
A poly learning rate policy \cite{chen2017deeplab} is used in our experiments. We set the initial learning rates as $0.001$
for PASCAL Context \cite{mottaghi2014role}, $0.002$ for PASCAL VOC 2012
\cite{everingham2010pascal} and 
ADE20K \cite{zhou2017scene}. The power of poly learning rate policy is set as $0.9$. The optimizer
is stochastic gradient descent (SGD) \cite{bottou2010large} with momentum $0.9$ and weight
decay $0.0001$.
We train our EfficientFCN for $120$ epochs on PASCAL Context, $80$ epochs on PASCAL 2012 and $120$ epochs on ADE20K, respectively. We set the crop size to $512\times 512$ on PASCAL Context and PASCAL 2012. Since the average image size is larger than other two datasets, we use $576 \times 576$ as the crop size on ADE20K. For data augmentation, we only randomly flip the input image and scale it randomly in the range $[0.5, 2.0]$.
%
%

\noindent
\textbf{Evaluation metrics.}\
We choose the standard evaluation metrics of pixel accuracy (pixAcc) and mean Intersection of Union (mIoU) as the evaluation metrics in our experiments. Following the best practice \cite{Zhang_2018_CVPR,he2019adaptive,fu2019dual}, 
we apply the strategy of averaging the network predictions in multiple scales for evaluation. For
each input image, we first randomly resize the input image with a scaling factor sampled uniformly
from [0.5, 2.0] and also randomly horizontally flip the image. These predictions are then averaged to generate the final prediction.

\subsubsection{Results on PASCAL Context}
The PASCAL Context dataset consists of 4,998 training images and 5,105 testing images for scene parsing. It is a complex and challenging dataset based on PASCAL VOC 2010 with more annotations and fine-grained scene classes, which includes 59 foreground classes and one background class. 
We take the same experimental settings and evaluation strategies following previous works \cite{Zhang_2018_CVPR,Zhang_2019_CVPR,fu2019dual,Fu_2019_ICCV,he2019dynamic}.
We first conduct ablation studies on this dataset to demonstrate the
effectiveness of each individual module design of our proposed EfficientFCN
and then compare our model with state-of-the-art methods. The ablation studies are conducted with a ResNet101 encoder backbone.

\begin{table}[tb]
\centering
\begin{minipage}[t]{0.5\textwidth}
%
\centering
\caption{Results of using different numbers of scales for multi-scale fused feature $m_{32}$ to generate the holistic codewords.}
\label{table:ablation_ms_words}
\setlength{\tabcolsep}{1mm}{
\begin{tabular}{c|cccc}
\hline
        {} & {$\{32\}$} & {$\{16, 32\}$} & {$\{8, 16, 32\}$} \\ \hline\hline
        pixAcc &80.0 & 80.1 & 80.3  \\
        mIoU & 54.8 & 55.1 & 55.3   \\
\hline
\end{tabular}}
\end{minipage}
\end{table}

\begin{table}[t]
\begin{minipage}[t]{0.5\textwidth}
\centering
\caption{Results of using different numbers of scales for multi-scale fused feature $m_8$ to estimate codeword assembly coefficients.}
\label{table:ablation_ms_assembly}
\setlength{\tabcolsep}{1mm}{
\begin{tabular}{c|cccc}
\hline
        {} & {$\{8\}$} & {$\{8, 16\}$} & {$\{8,16,32\}$} \\ \hline\hline
        pixAcc &78.9 & 80.0 & 80.3  \\
        mIoU & 47.9 & 52.1 & 55.3   \\
\hline
\end{tabular}}
\end{minipage}
\end{table}

\noindent \textbf{Comparison with the classical encoder-decoders.} For the classical encoder-decoder based methods, the feature upsampling is achieved via either bilinear interpolation or deconvolution. We implement two classical encoder-decoder based methods, which include the
feature upsampling operation (bilinear upsampling or deconvolution) and the skip-connections. 
To verify the effectiveness of our proposed HGD,  these two methods are trained and tested 
on the PASCAL Context dataset with the same training setting as our model.
The results are shown in Table \ref{table:ablation_encoder_decoder}. Although the classical 
encoder-decoder methods have similar computational complexities, their performances 
are generally far inferior than our EfficientFCN. The key reason is that their upsampled 
feature maps are created in a local manner. The simple bilinear interpolation or 
deconvolution cannot effectively upsample the OS=32 feature maps even with the skip-connected 
OS=8 and OS=16 feature maps. In contrast, our proposed HGD can effectively upsample the 
high-resolution semantic-rich feature maps not only based on the fine-grained structural 
information in the OS=8 and OS=16 feature maps but also from the holistic semantic information from the OS=32 feature maps.  

\noindent \textbf{Multi-scale feature fusion.} We conduct two experiments on multi-scale features fusion to verify their effects on holistic codeword generation and codeword assembly for feature upsampling. In our holistically-guided decoder, the holistic codewords are generated based on the OS=32 multi-scale fused feature maps $m_{32}$ and the codeword assembly coefficients are predicted from the OS=8 multi-scale fused feature maps $m_8$.
For the codeword generation, we conduct three experiments to generate the holistic codewords from
multi-scale fused features with different numbers of scales. As shown in Table
\ref{table:ablation_ms_words}, when reducing the number of fusion scales from 3 to 2 and 
from 2 to 1, the performances of our EfficientFCN slightly decrease. The phenomenon is reasonable as the deepest feature maps contain more categorical information than the OS=8 and OS=16 feature maps. 
For the codeword assembly coefficient estimation, the similar experiments are conducted, where results are shown in Table \ref{table:ablation_ms_assembly}. However, different from the above results, when fewer scales of feature maps are used to form the multi-scale fused OS=8 feature map $m_8$, the performances of our EfficientFCN show significant drops.
These results demonstrate that although the OS=8 feature maps contain more fine-grained structural information, the semantic features from higher-level feature maps are essential for guiding the recovery of the semantic-rich high-resolution feature maps.
\begin{table}[!t]
\centering
%
\caption{Ablation study of the number of the holistic codewords.}
\label{table:ablation_n_codewords}
\setlength{\tabcolsep}{2mm}{
\begin{tabular}{c|cccccc}
\hline
        {} & {32} & {64} & {128}  & {256} &{512} &{1024}\\ \hline\hline
        pixAcc &79.9 & 80.1 & 80.1 & 80.3 & 80.3 &80.1 \\
        mIoU & 54.5 & 54.9 & 55.0 &55.3 & 55.5 & 55.1  \\
        GFLOPS & 67.9 & 68.1 & 68.6 & 69.6 & 72.1 & 78.9 \\
\hline
\end{tabular}}
\end{table}

\begin{table*}[!t]
\centering
\caption{Segmentation results of state-of-the-art methods on PASCAL Context and ADE20K validation dataset.}
\label{table:pascal-context}
\centering
\begin{tabular}{lllll}
\hline
    \textbf{Method} & \textbf{Backbone} &\begin{tabular}{c} 
        \textbf{mIoU\%}\\ (PASCAL Context) \end{tabular} & \begin{tabular}{c} 
        \textbf{mIoU\%} \\ (ADE20K) \end{tabular} & \textbf{GFlops} \\\hline\hline
    DeepLab-v2 \cite{chen2017deeplab} &  Dilated-ResNet101-COCO & 45.7 & - & $>$223\\
    RefineNet \cite{RefineNet} &  Dilated-ResNet152 & 47.3 & - & $>$223\\
    MSCI \cite{MSCI} &  Dilated-ResNet152 & 50.3 & - & $>$223\\
    PSPNet \cite{zhao2017pyramid} & Dilated-ResNet101 & - & 43.29 & $>$223 \\
    SAC \cite{zhang2017scale} & Dilated-ResNet101 & - & 44.30 & $>$223  \\
    EncNet \cite{Zhang_2018_CVPR} &  Dilated-ResNet101 & 51.7 & 44.65 & 234\\
    DANet \cite{fu2019dual} &  Dilated-ResNet101 & 52.6 & - & $>$223 \\ 
    APCNet \cite{he2019adaptive} &  Dilated-ResNet101 & 54.7 & 45.38 & 245\\ 
    CFNet \cite{Zhang_2019_CVPR} &  Dilated-ResNet101 & 54.0 & 44.89 & $>$223 \\ 
    ACNet \cite{Fu_2019_ICCV} & Dilated-ResNet101 & 54.1 & \textbf{45.90} & $>$223\\ 
    APNB \cite{zhu2019asymmetric} & Dilated-ResNet101 & 52.8 & 45.24 & $>$223  \\ 
    DMNet \cite{he2019dynamic} & Dilated-ResNet101 & 54.4 & 45.50 & 242\\
\hline
    Ours & ResNet101 & \textbf{55.3} & {45.28} & \textbf{70} \\ \hline
\end{tabular}
\end{table*}

\begin{table*}[!t]
\small
\centering
\caption{Results of each category on PASCAL VOC 2012 test set. Our
        EfficientFCN obtains 85.4 \% without MS COCO dataset pre-training and 87.6\% with MS COCO dataset pre-training. (For each column, the best two entries are masked in shaded regions.)
    }
\label{table:pascal_voc_2012}
\resizebox{\textwidth}{!}{%
\begin{tabular}{l|cccccccccccccccccccc|c}
\hline
\textbf{Method}    & \textbf{aero} & \textbf{bike} & \textbf{bird} & \textbf{boat} & \textbf{bottle} & \textbf{bus}  & \textbf{car}  & \textbf{cat}  & \textbf{chair} & \textbf{cow}  & \textbf{table} & \textbf{dog}  & \textbf{horse} & \textbf{mbike} & \textbf{person} & \textbf{plant} & \textbf{sheep} & \textbf{sofa} & \textbf{train} & \textbf{tv}   & \textbf{mIoU\%} \\ \hline\hline
\textbf{FCN} \cite{long2015fully}       & 76.8          & 34.2          & 68.9
& 49.4          & 60.3            & 75.3          & 74.7          & 77.6
& 21.4           & 62.5          & 46.8           & 71.8          & 63.9
& 76.5           & 73.9            & 45.2           & 72.4           & 37.4
& 70.9           & 55.1          & 62.2   \\
\textbf{DeepLabv2} \cite{chen2017deeplab} & 84.4          & 54.5          & 81.5
& 63.6          & 65.9            & 85.1          & 79.1          & 83.4
& 30.7           & 74.1          & 59.8           & 79.0            & 76.1           & 83.2           & 80.8            & 59.7           & 82.2           & 50.4          & 73.1           & 63.7          & 71.6  \\
\textbf{CRF-RNN} \cite{CRF-RNN}   & 87.5          & 39.0          & 79.7          & 64.2          & 68.3            & 87.6          & 80.8          & 84.4          & 30.4           & 78.2          & 60.4           & 80.5          & 77.8           & 83.1           & 80.6            & 59.5           & 82.8           & 47.8          & 78.3           & 67.1          & 72.0  \\
\textbf{DeconvNet} \cite{DeconvNet} & 89.9          & 39.3          & 79.7          & 63.9          & 68.2            & 87.4          & 81.2          & 86.1          & 28.5           & 77.0          & 62.0           & 79.0          & 80.3           & 83.6           & 80.2            & 58.8           & 83.4           & 54.3          & 80.7           & 65.0            & 72.5   \\
\textbf{DPN} \cite{DPN}       & 87.7          & 59.4          & 78.4          & 64.9          & 70.3            & 89.3          & 83.5          & 86.1          & 31.7           & 79.9          & 62.6           & 81.9          & 80.0           & 83.5           & 82.3            & 60.5           & 83.2           & 53.4          & 77.9           & 65.0            & 74.1   \\
\textbf{Piecewise} \cite{Piecewise} & 90.6          & 37.6          & 80.0
& 67.8          & 74.4            & 92            & 85.2          & 86.2
& 39.1           & 81.2          & 58.9           & 83.8          & 83.9
& 84.3           & 84.8            & 62.1           & 83.2           & 58.2
& 80.8           & 72.3          & 75.3    \\
\textbf{ResNet38} \cite{ResNet38}  & 94.4          & 72.9          & 94.9
& 68.8          & 78.4            & 90.6          & 90.0          & 92.1
& 40.1           & 90.4          & 71.7           & 89.9          & 93.7
& \bgGray 91.0           & 89.1            & 71.3           & 90.7           & 61.3          & 87.7           & 78.1          & 82.5     \\
\textbf{PSPNet} \cite{zhao2017pyramid}    & 91.8          & 71.9          & 94.7          & 71.2          & 75.8            & 95.2          & 89.9          & 95.9          & 39.3           & 90.7          & 71.7           & 90.5          & 94.5           & 88.8           & 89.6            & 72.8           & 89.6           & \bgGray {64.0}          & 85.1           & 76.3          & 82.6   \\
\textbf{EncNet} \cite{Zhang_2018_CVPR}    & 94.1          & 69.2          & \bgGray\textbf{96.3} & \bgGray 76.7          & \bgGray \textbf{86.2}   & 96.3          & 90.7          & 94.2          & 38.8           & 90.7          & 73.3           & 90.0          & 92.5           & 88.8           & 87.9            & 68.7           & 92.6           & 59.0          & 86.4           & 73.4          & 82.9            
            \\
            \textbf{APCNet} \cite{he2019adaptive}      & 95.8 &\bgGray 75.8 & 84.5  & 76.0 & 80.6 &
            \bgGray 96.9 & 90.0 & 96.0 & \bgGray\textbf{42.0} & \bgGray 93.7
            &\bgGray 75.4 & 91.6 & 95.0 & 90.5 &
             89.3 &  75.8 & 92.8 & 61.9 &  88.9 & \bgGray 79.6 & 84.2
            \\
            \textbf{CFNet} \cite{Zhang_2019_CVPR}      & 95.7 & 71.9 &\bgGray
            95.0  &\bgGray 76.3 & \bgGray 82.8 &
            94.8 & 90.0 & 95.9 & 37.1 & 92.6 & 73.0 & \bgGray 93.4 & 94.6 & 89.6 &
            88.4 & 74.9 & \bgGray \textbf{95.2} &  63.2 & \bgGray \textbf{89.7} & 78.2 & 84.2
            \\
            \textbf{DMNet} \cite{he2019dynamic}        & \bgGray 96.1 &
            \bgGray\textbf{77.3} & 94.1 & 72.8 & 78.1 & 
            \bgGray\textbf{97.1} & \bgGray \textbf{92.7} & \bgGray 96.4 & 39.8 & 91.4 & \bgGray 75.5 & 92.7 & \textbf{95.8} &
            \bgGray {91.0} & \bgGray {90.3} & \bgGray {76.6} & \bgGray 94.1 & 62.1 & 85.5 & 77.6 & \bgGray 84.4 
            \\ \hline
            \textbf{Ours}      & \bgGray \textbf{96.4} & {74.1} &
            92.8 & \bgGray 75.6 & 81.9 &\bgGray 96.9 
            & \bgGray {92.6}  & \bgGray \textbf{97.1} & \bgGray 41.6 & \bgGray \textbf{95.4} 
            & 72.9 & \bgGray \textbf{93.9} & \bgGray \textbf{95.9} 
            & {90.6} &\bgGray \textbf{ 90.6} &\bgGray \textbf{77.2}  & 94.0 &
            67.5 &\bgGray 89.3 &
            \bgGray \textbf{79.8} & \bgGray \textbf{85.4} \\
             \hline
            \multicolumn{22}{c}{\textbf{With COCO Pre-training}}\\
            \hline
            
            \textbf{CRF-RNN}~\cite{CRF-RNN} & 90.4 & 55.3 & 88.7 & 68.4 & 69.8 & 88.3 & 82.4 & 85.1 & 32.6 & 78.5 & 64.4 & 79.6 & 81.9 & 86.4 & 81.8 & 58.6 & 82.4 & 53.5 & 77.4 & 70.1 & 74.7 \\
            \textbf{Piecewise}~\cite{Piecewise} & 94.1 & 40.7 & 84.1 & 67.8 & 75.9 & 93.4 & 84.3 & 88.4 & 42.5 & 86.4 & 64.7 & 85.4 & 89.0 & 85.8 & 86.0 & 67.5 & 90.2 & 63.8 & 80.9 & 73.0 & 78.0 \\
            \textbf{DeepLabv2}~\cite{chen2017deeplab} & 92.6 & 60.4 & 91.6 & 63.4 & 76.3 & 95.0 & 88.4 & 92.6 & 32.7 & 88.5 & 67.6 & 89.6 & 92.1 & 87.0 & 87.4 & 63.3 & 88.3 & 60.0 & 86.8 & 74.5 & 79.7  \\
            \textbf{RefineNet}\cite{RefineNet}  & 95.0 & 73.2 & 93.5 & 78.1 & 84.8 & 95.6 & 89.8 & 94.1 & 43.7 & 92.0 & 77.2 & 90.8 & 93.4 & 88.6 & 88.1 & 70.1 & 92.9 & 64.3 & 87.7 & 78.8 & 84.2 \\
            \textbf{ResNet38}\cite{ResNet38}  &  96.2 & 75.2 & \secbest
            95.4 & 74.4 & 81.7 & 93.7 & 89.9 & 92.5 & \secbest 48.2 & 92.0 & 79.9
            & 90.1 & 95.5 & 91.8 & 91.2 &  73.0 & 90.5 & 65.4 & 88.7 & 80.6 & 84.9\\
            \textbf{PSPNet}~\cite{zhao2017pyramid} & {95.8} & {72.7} & {95.0}
            & {78.9} & {84.4} & 94.7 & \secbest{92.0} & {95.7} & {43.1} &
            {91.0} & \best{80.3} & {91.3} & {96.3} & {92.3} & {90.1} & {71.5}
            & \secbest{94.4} & \secbest{66.9} & {88.8} & \best{82.0} & {85.4} \\
            \textbf{DeepLabv3}\cite{chen2017rethinking} & \secbest 96.4 & 76.6
            & 92.7 & 77.8 & \secbest{87.6} & 96.7 & 90.2 & 95.4 &  47.5 &
            \secbest 93.4 & 76.3 & 91.4 & \best{97.2} &  91.0 & \best{92.1} &
            71.3 & 90.9 & \secbest{68.9} & \secbest{90.8} & 79.3 & 85.7 \\ 
            \textbf{EncNet}\cite{Zhang_2018_CVPR}  & 95.3 & 76.9 & 94.2 &
            \secbest 80.2 & 85.2 & 96.5 & 90.8 & 96.3 &  47.9
            &  93.9 & \secbest 80.0 & 92.4 & \secbest 96.6 & 90.5 &  91.5 & 70.8 &  93.6 & 66.5 & 87.7 & 80.8 & 85.9 
            \\ 
            \textbf{CFNet} \cite{Zhang_2019_CVPR}      &\best 96.7 &\secbest 79.7 &
            94.3  & 78.4 & 83.0 & \best 97.7 & 91.6 &\secbest 96.7 &\best 50.1
            &\secbest 95.3 & 79.6 & \bgGray 93.6 &\best 97.2 &\secbest
            94.2 &\secbest 91.7 & \bgGray {78.4} &\best  95.4 & \bgGray
            \textbf{69.6} & 90.0 & 81.4 & \secbest 87.2
            \\ \hline
            \textbf{Ours} & \bgGray {96.6} & \best {80.6} &
            \best 96.1 & \best{82.3} &\best 87.8 &\best 97.7 
            & \best {94.4}  & \best {97.3} &  47.1 & \bgGray \textbf{96.3} 
            & {77.9} & \bgGray \textbf{94.8} & \bgGray
            \textbf{97.2} 
            &\bgGray \textbf{94.3} & 91.1 & \best 81.0  & 94.3 & 61.5
            &\best 91.6 &\best {83.5} & \bgGray \textbf{87.6}
            \\ \hline
\end{tabular}}
\end{table*}

\begin{figure*}[!t]
\centering
\begin{center}
\begin{tabular}{C{2.5cm}C{2.5cm}C{2.5cm}C{2.5cm}C{2.5cm}}
    \includegraphics[width=2.50cm]{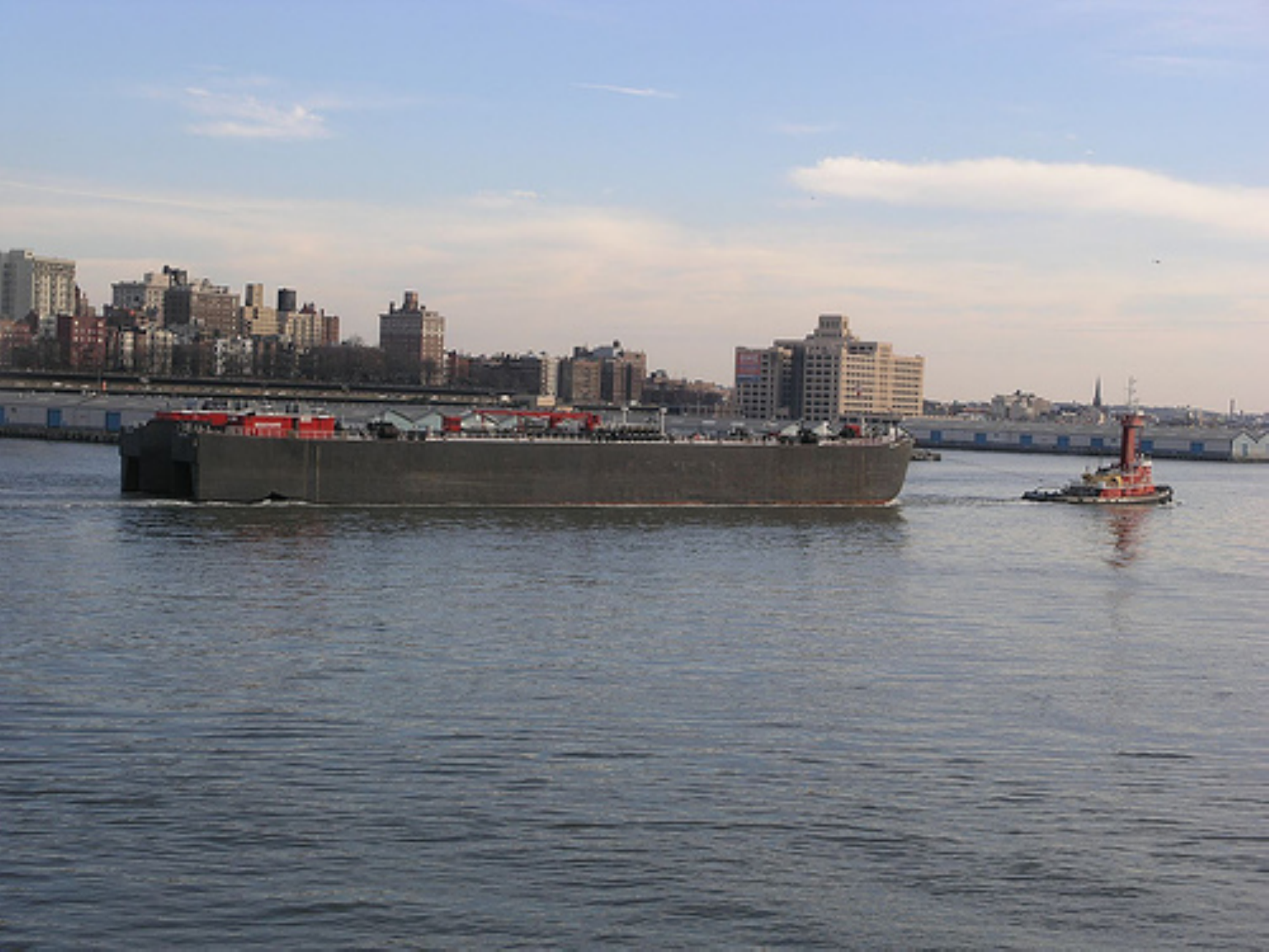} &
    \includegraphics[width=2.50cm]{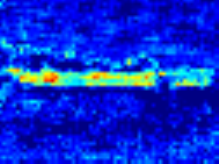} &
    \includegraphics[width=2.50cm]{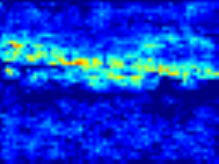} &
    \includegraphics[width=2.50cm]{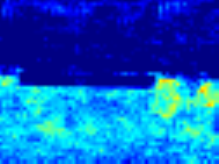} &
    \includegraphics[width=2.50cm]{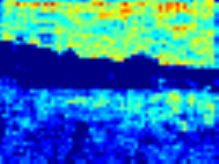} \\
    \includegraphics[width=2.50cm]{figs/exp1_fig1_a} &
    \includegraphics[width=2.50cm]{figs/exp1_fig1_b} &
    \includegraphics[width=2.50cm]{figs/exp1_fig1_c} &
    \includegraphics[width=2.50cm]{figs/exp1_fig1_d} &
    \includegraphics[width=2.50cm]{figs/exp1_fig1_e} \\
    \includegraphics[width=2.50cm]{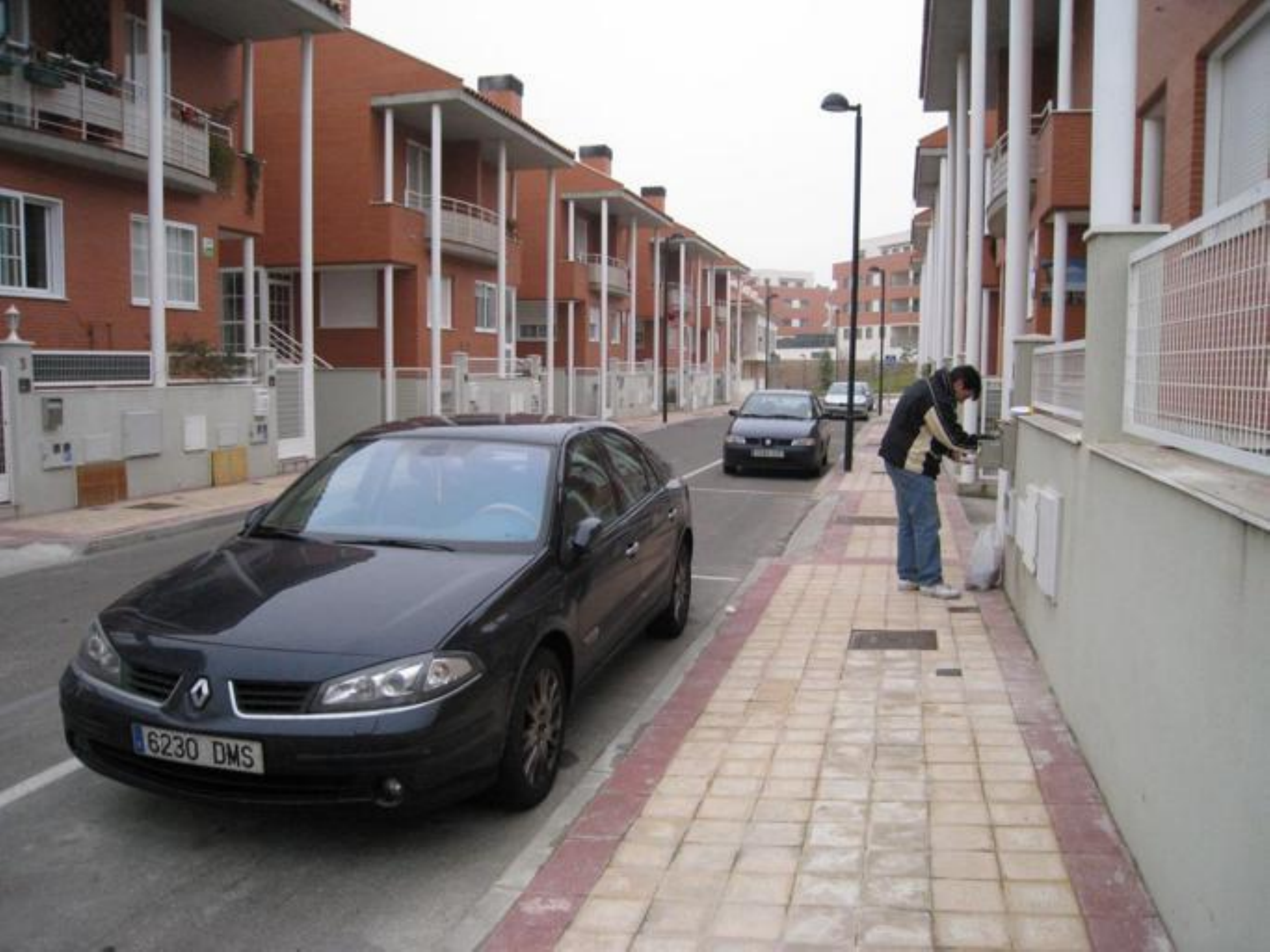} &
    \includegraphics[width=2.50cm]{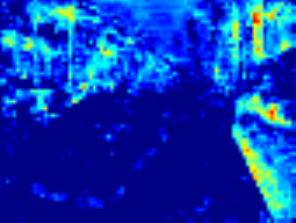} &
    \includegraphics[width=2.50cm]{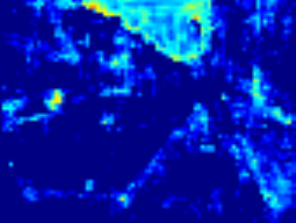} &
    \includegraphics[width=2.50cm]{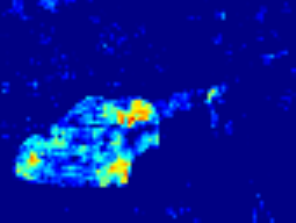} &
    \includegraphics[width=2.50cm]{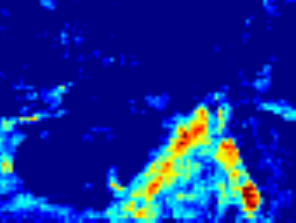} \\
    \centering (a) & (b) & (c) & (d) & (e) \\
\end{tabular}
\end{center}
\caption{(a) Input images from the PASCAL Context and ADE20K dataset. (b-e) Different weighting maps $\tilde{A}_i$ for creating the holistic codewords.}
\label{fig:weighting_maps}
\end{figure*}

\begin{figure*}[tb]
\begin{center}
\begin{tabular}{C{2.90cm}C{2.90cm}C{2.90cm}C{2.90cm}}
    \includegraphics[width=2.90cm]{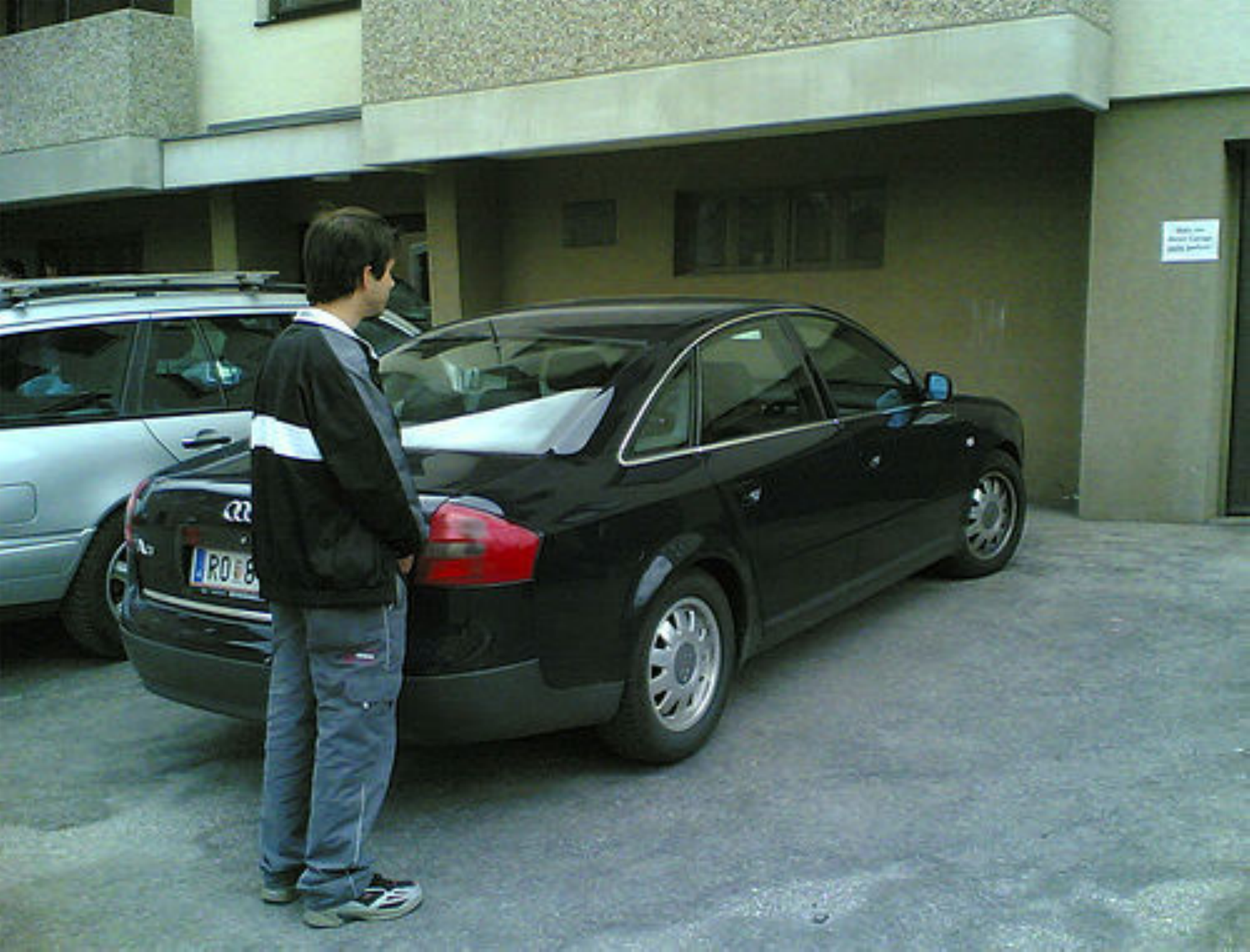} &
    \includegraphics[width=2.90cm]{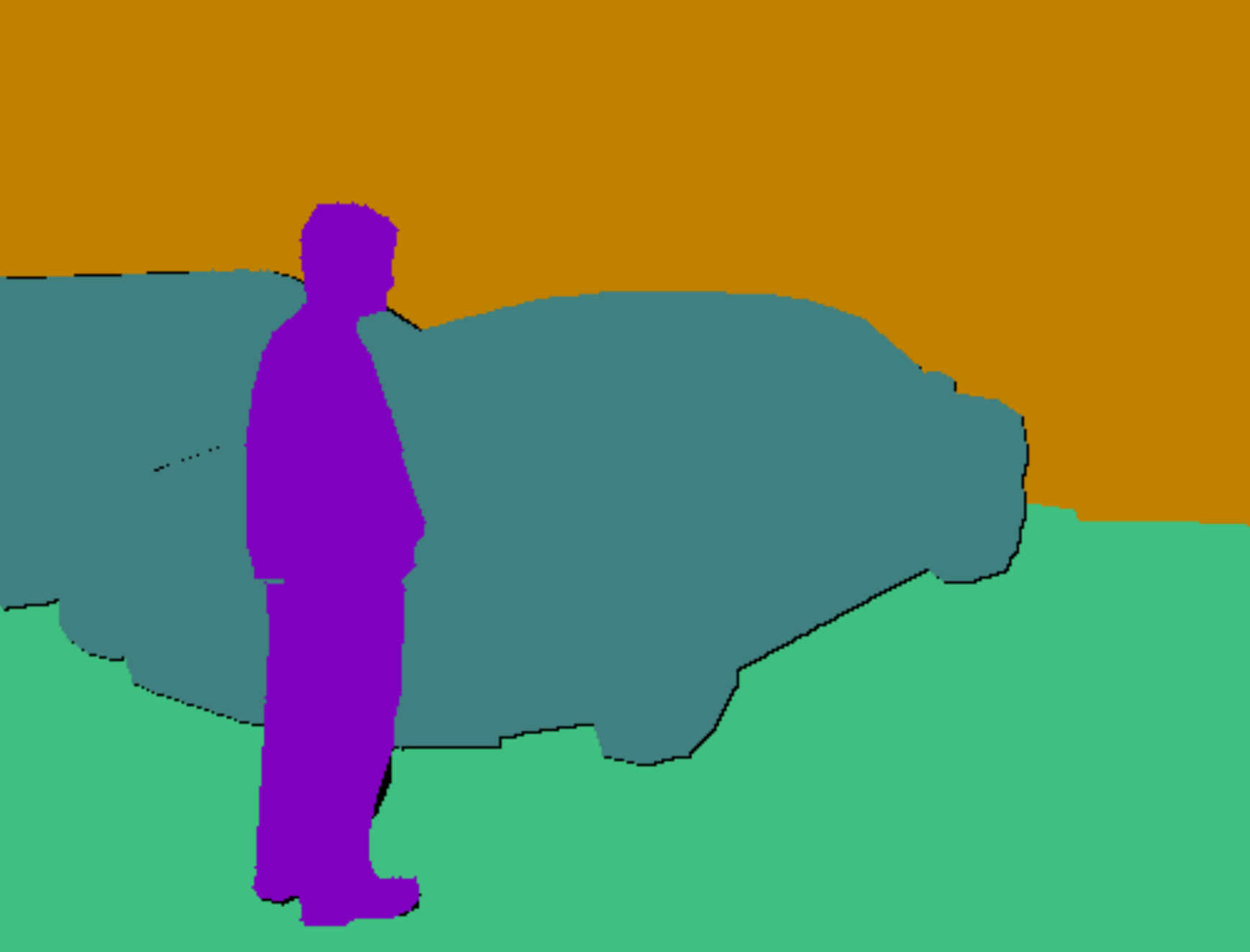} &
    \includegraphics[width=2.90cm]{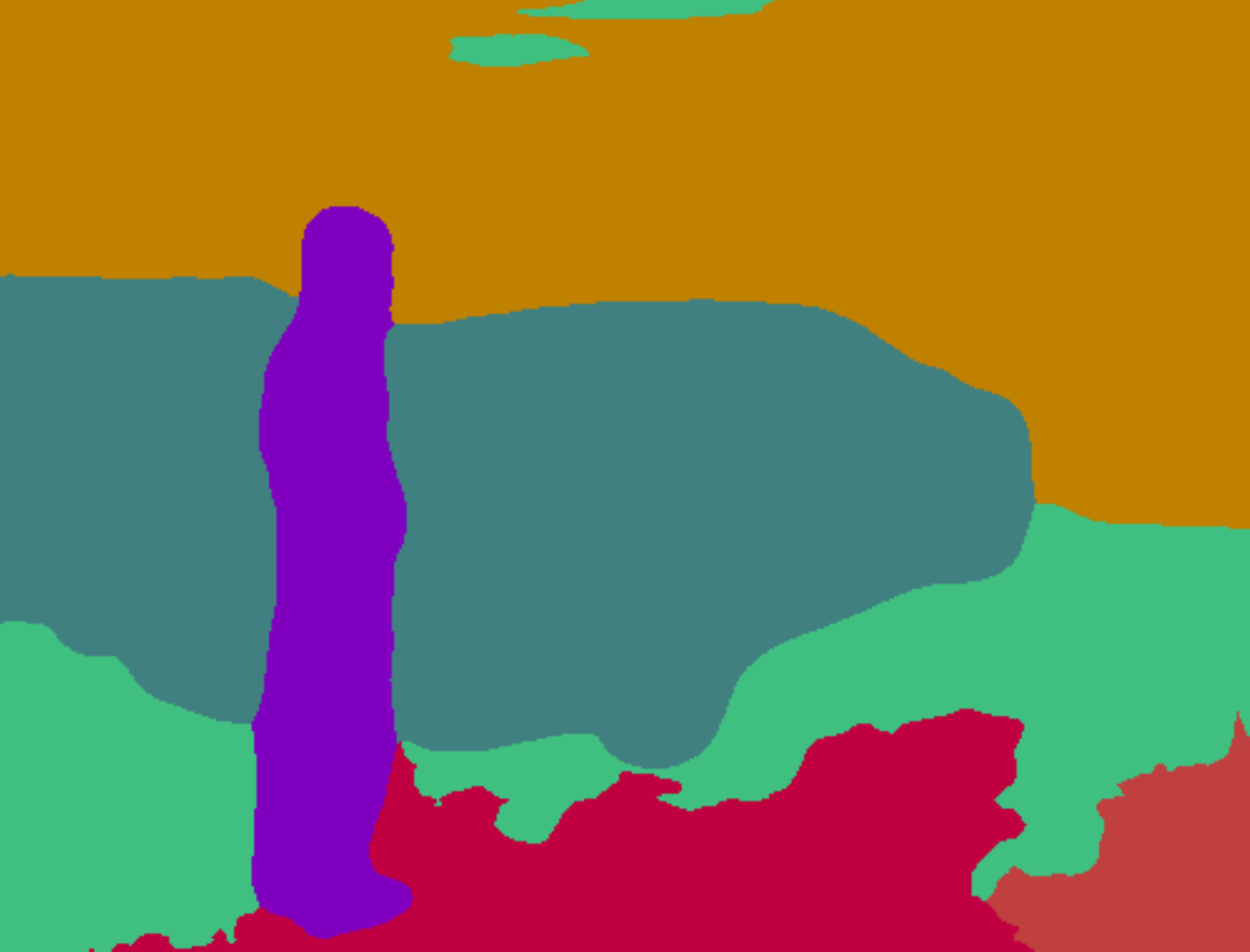} &
    \includegraphics[width=2.90cm]{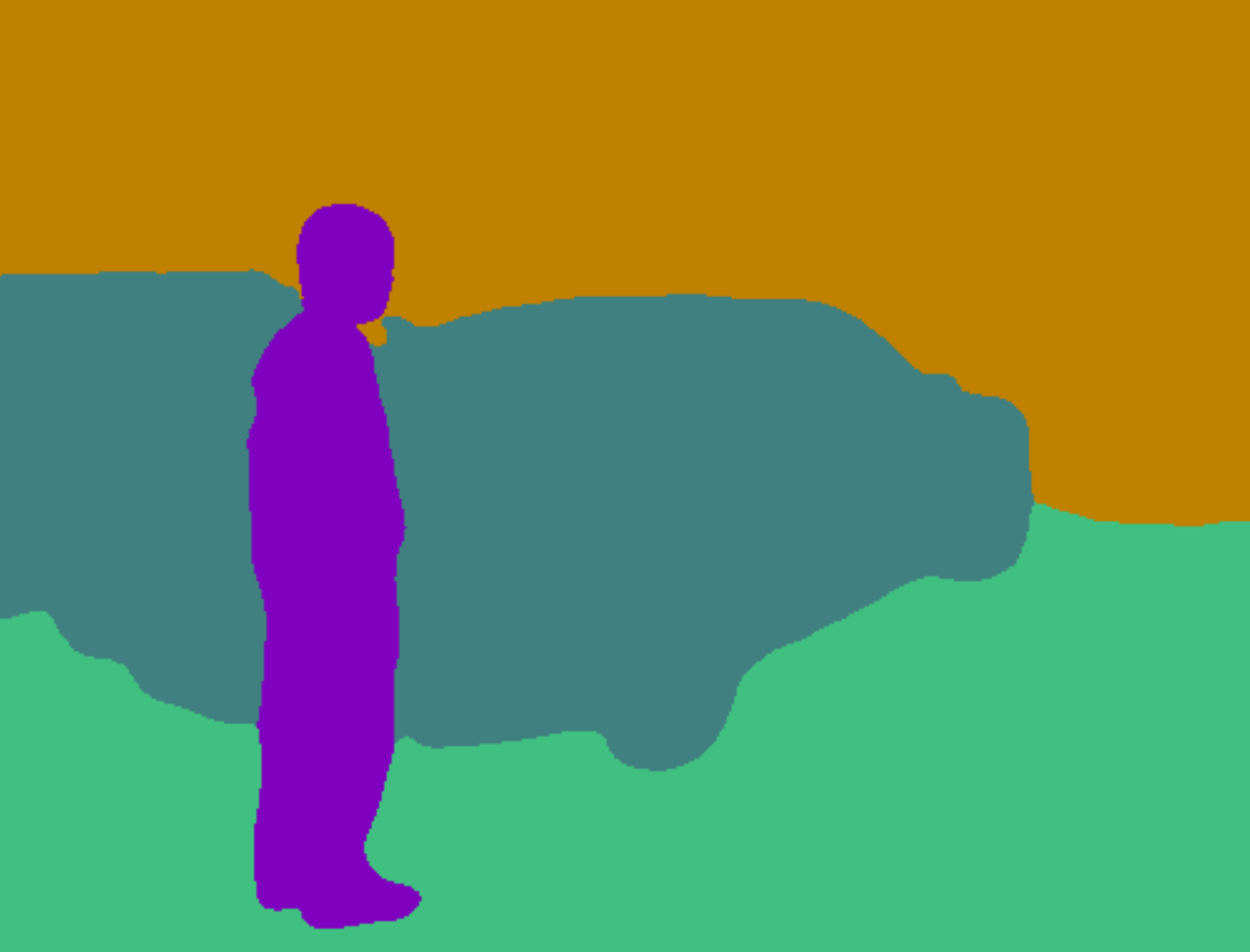} \\
    \includegraphics[width=2.90cm]{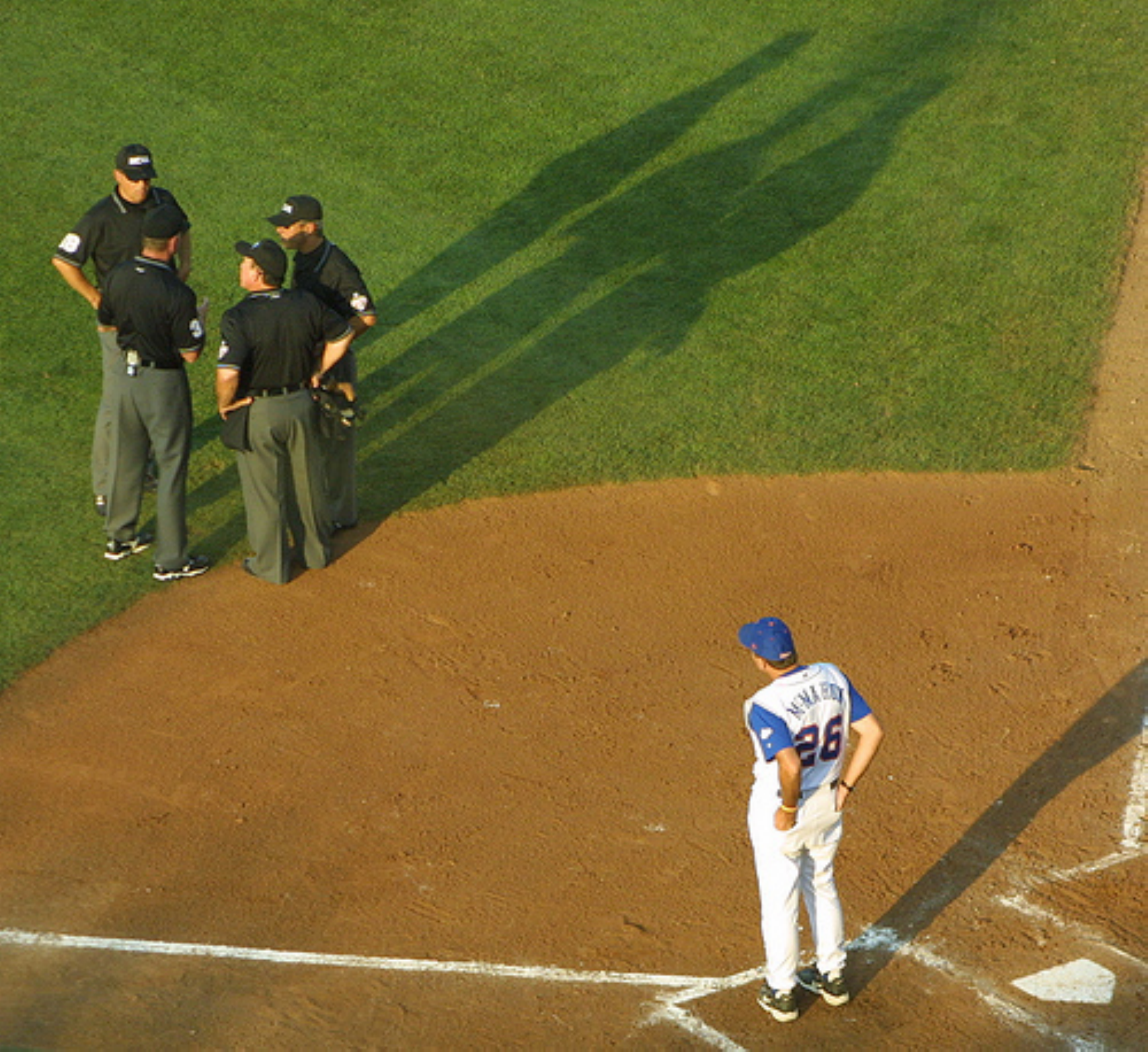} &
    \includegraphics[width=2.90cm]{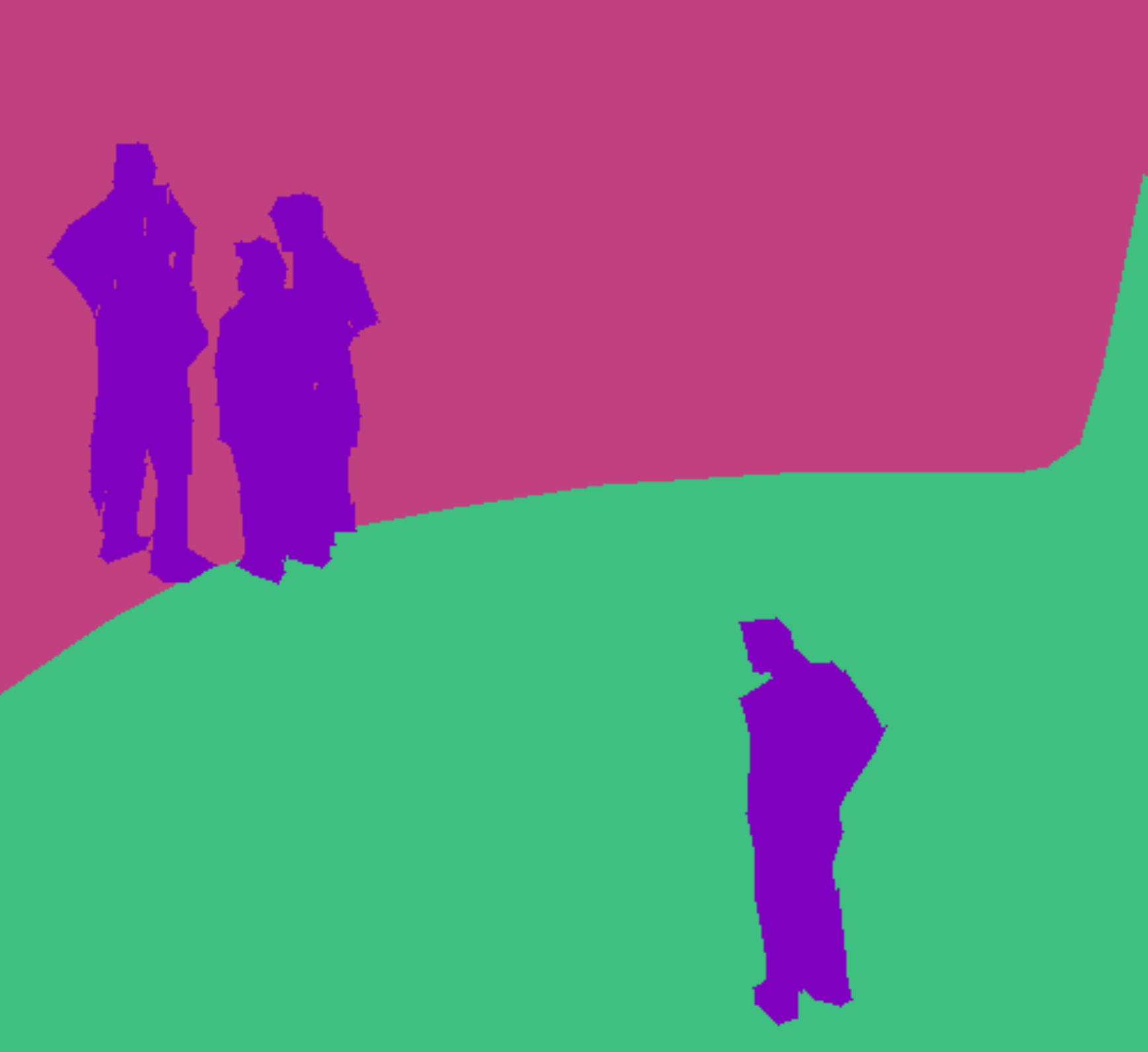} &
    \includegraphics[width=2.90cm]{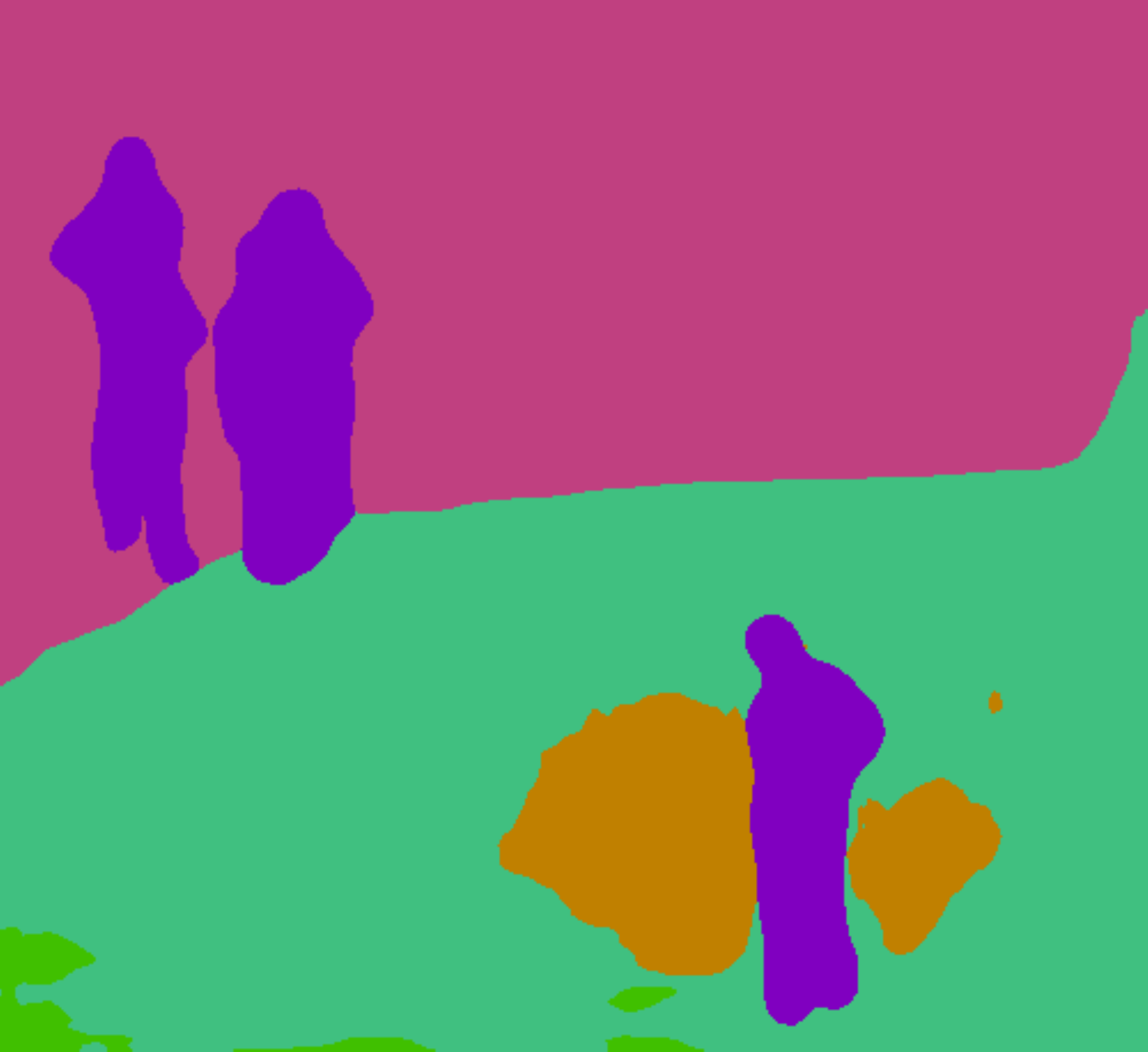} &
    \includegraphics[width=2.90cm]{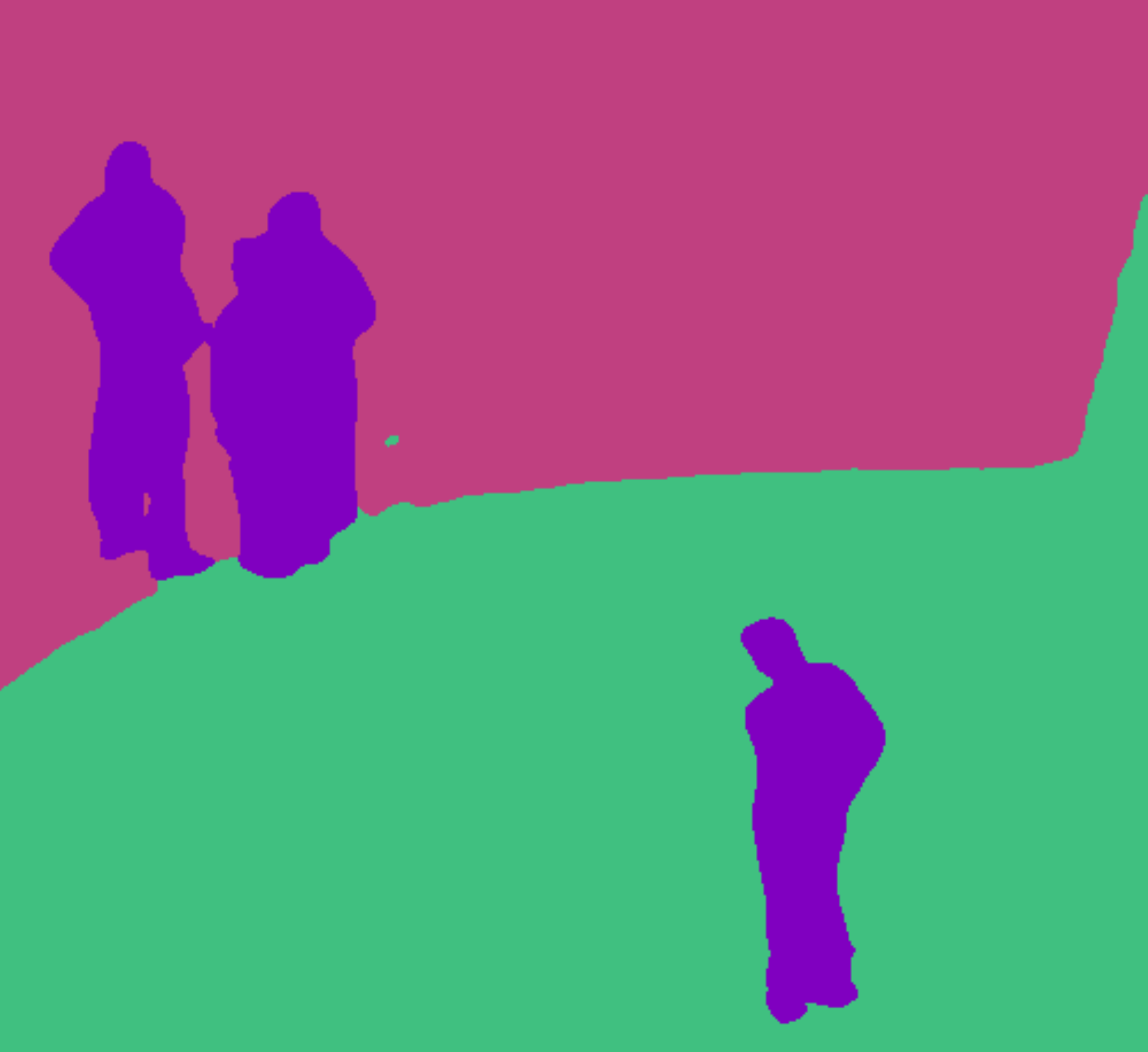} \\
    \includegraphics[width=2.90cm]{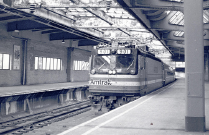} &
    \includegraphics[width=2.90cm]{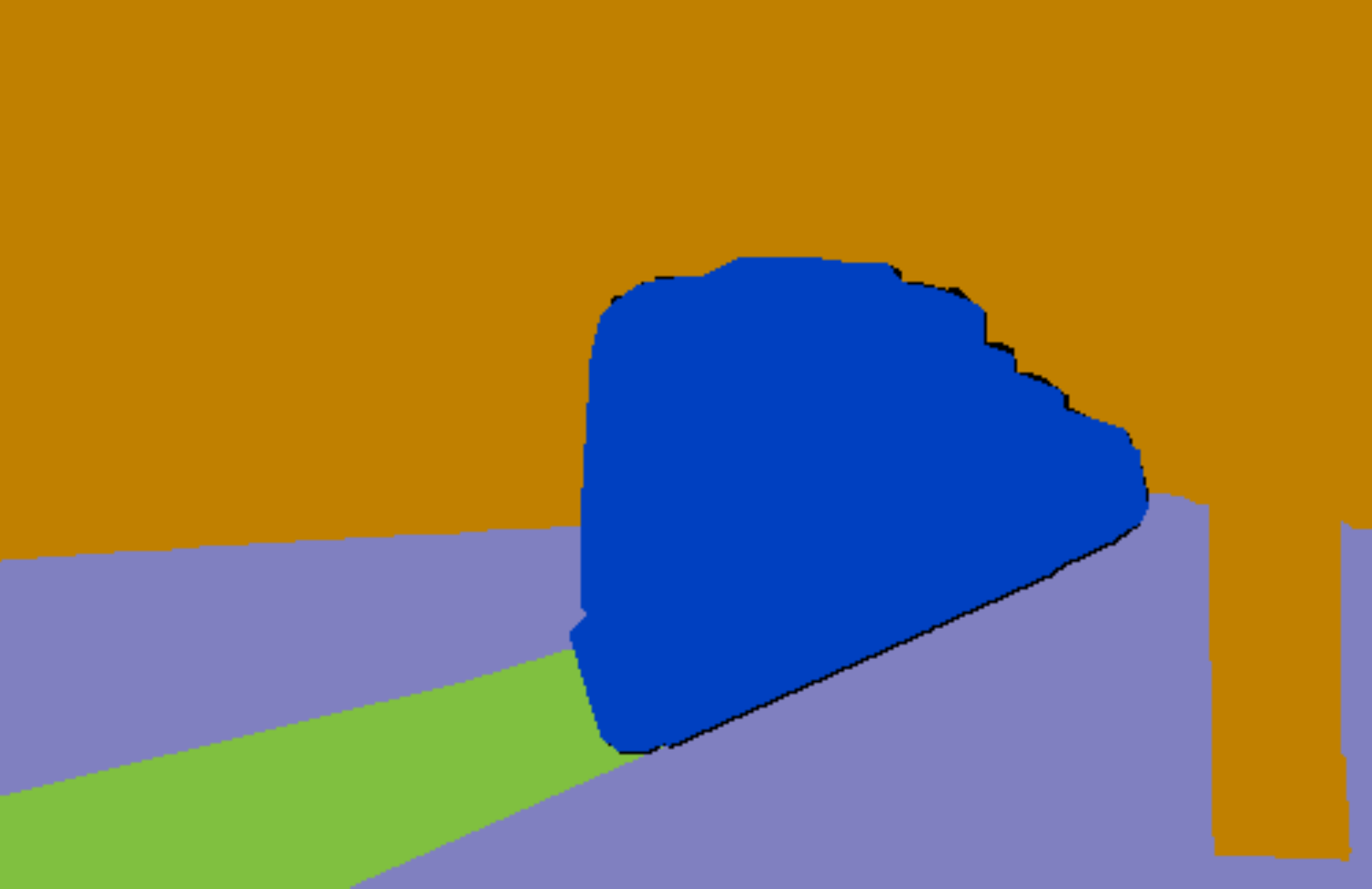} &
    \includegraphics[width=2.90cm]{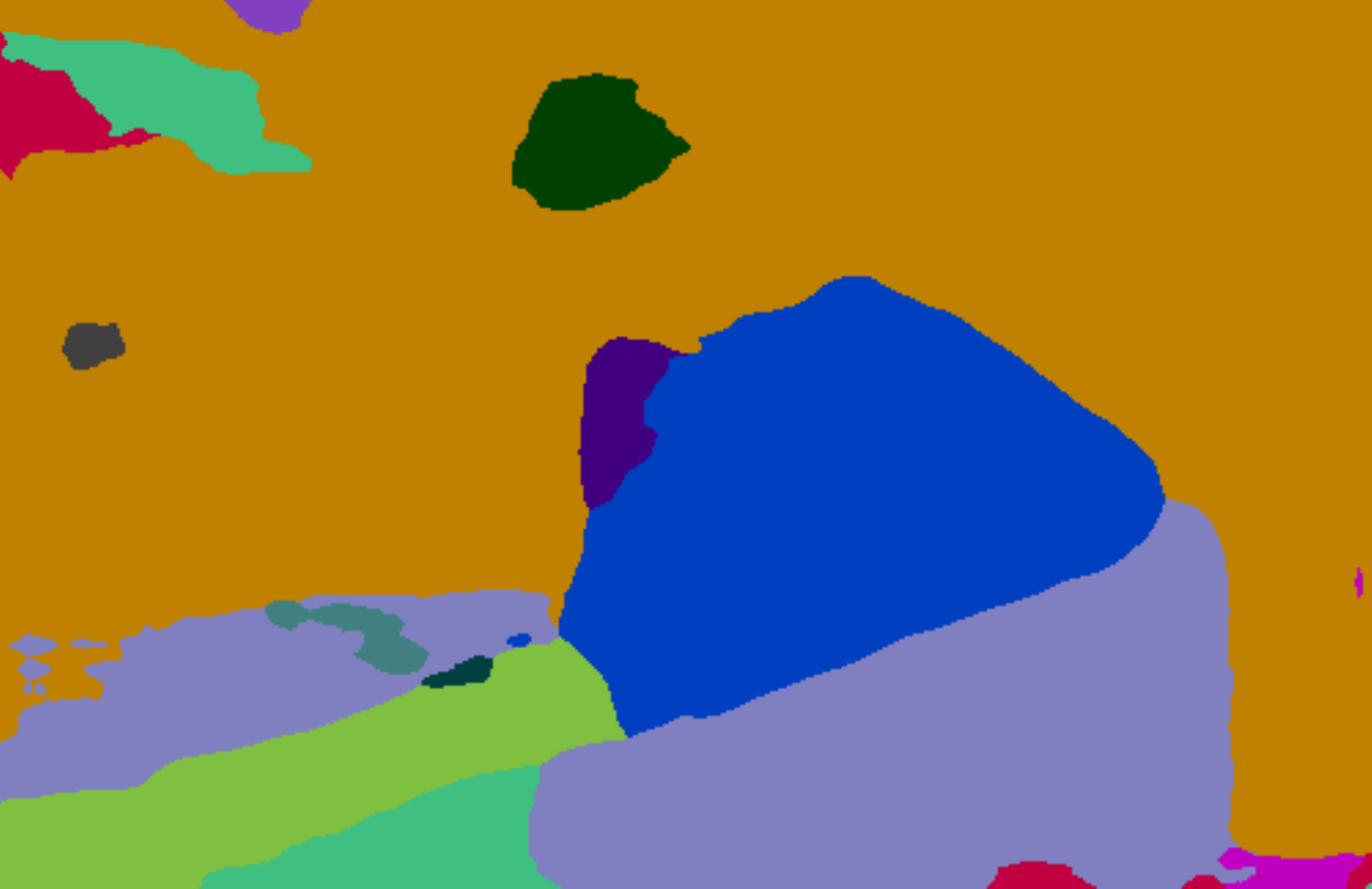} &
    \includegraphics[width=2.90cm]{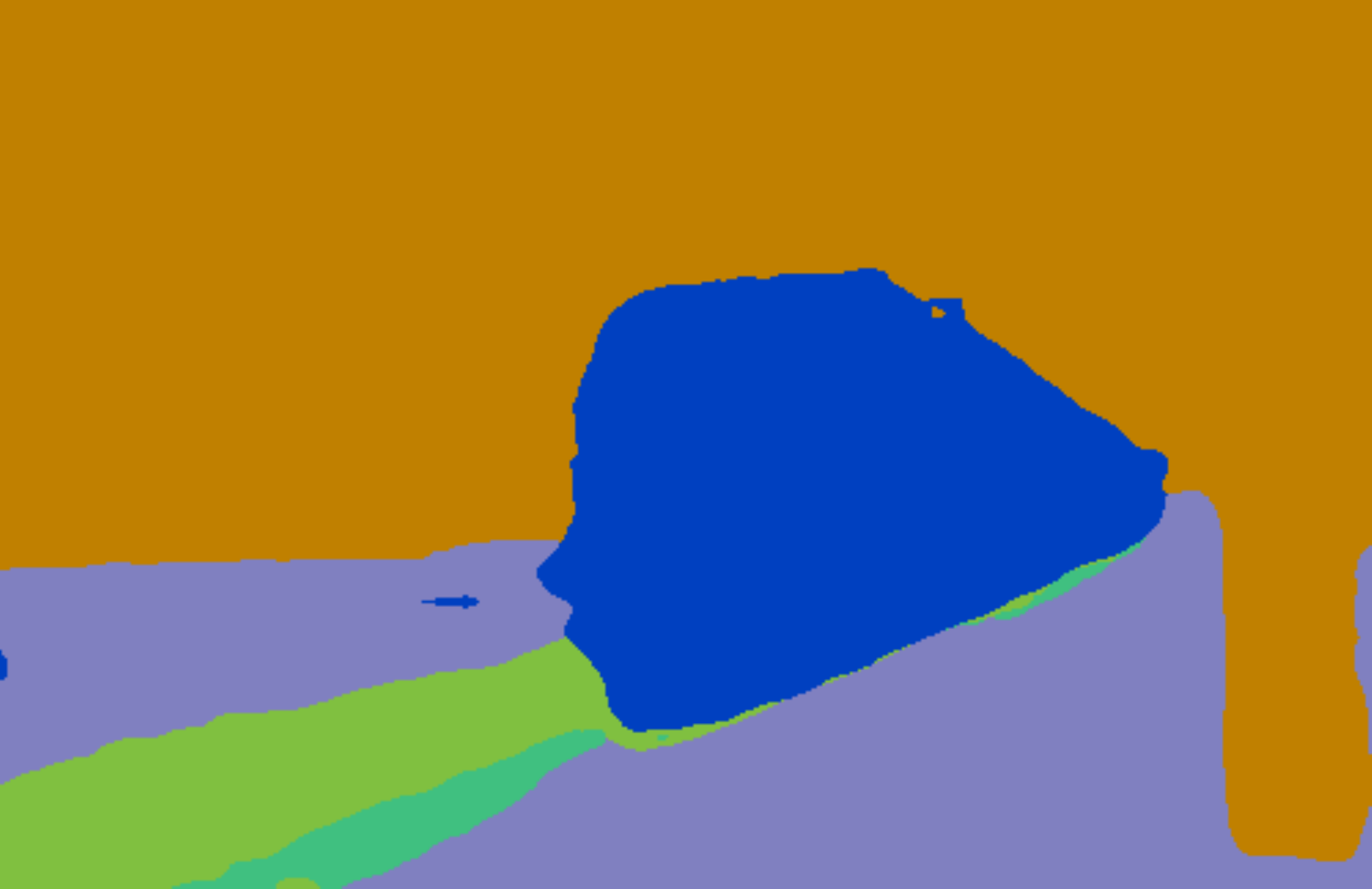} \\
    \centering (a) Image & (b) GT & (c) Baseline & (d) EfficientFCN \\
\end{tabular}
\end{center}
\caption{Visualization results from the PASCAL Context dataset.}
\label{fig:vis}
\end{figure*}

\noindent \textbf{Number of holistic codewords.} We also conduct  experiments to survey the
effectiveness of the number of codewords in our predicted holistic codewords for feature upsampling.
As shown in Table \ref{table:ablation_n_codewords}, as the number of the holistic codewords
increases from 32 to 512, the performance improves 1\% in terms of mIoU on PASCAL Context.
However, when the number of the holistic codewords further increases from 512 to 1024, the performance
has a slight drop, which might be caused by the additional parameters. The larger model capacity
might cause model to overfit the training data. In addition, since the assembly coefficients of the
holistic codewords are predicted from the OS=8 multi-scale fused feature $m_8$, the increased number
of the holistic codewords also leads to significantly more extra computational cost. Thus, to balance
the performance and also the efficiency, we set the number of the holistic codewords as 256 for the
PASCAL Context and PASCAL VOC 2012 datasets. Since PASCAL Context only has 60 classes and we observe
the number of codewords needed is approximately 4 times than the number of classes. We therefore set
the number of codewords as 600 for ADE20K, which has 150 classes.

\noindent
\textbf{Comparison with state-of-the-art methods.} 
To further demonstrate the effectiveness of our proposed EfficientFCN with the holistically-guided decoder, the comparisons with state-of-the-art methods are shown in Table \ref{table:pascal-context}. The dilatedFCN based methods dominate the top performances for semantic segmentation. However, our work is still able to achieve the best results compared to the dilatedFCN based methods on the PASCAL Context validation set without using any dilated convolution and has significantly less computational cost. Because of the efficient design of our HGD, our EfficientFCN only has 1/3 of the computational cost of state-of-the-arts methods but can still achieve the best performance. 

\noindent \textbf{Importance of the codeword information transfer for accurate assembly coefficient estimation.} 
The key of our proposed HGD is how to linearly assemble holistic codewords at each spatial location
to form high-resolution upsampled feature maps based on the feature maps $m_8$. In our HGD, although
the OS=8 features have well maintained structural image information, we argue that directly using OS=8 features to predict codeword assembly coefficients are less effective since they have no information about the codewords. 
We propose to transfer the codeword information as the average codeword basis,
which is location-wisely added to the OS=8 feature maps. To verify this
argument, we design an experiment that removes the additive information
transfer, and only utilizes two $1\times 1$ convolutions with the same output
channels on the OS=8 feature maps $m_8$ for directly predicting assembly
coefficients. The mIoU of this implementation is 54.2\%, which has a clear performance drop if there
is no codeword information transfer from the codeword generation branch to the codeword coefficient prediction branch.

\noindent \textbf{Visualization of the weighting maps and example results.} 
To better interpret the obtained holistic codewords, we visualize the weighting maps $\tilde{A}$ for
creating the holistic codewords in Fig.~\ref{fig:weighting_maps}, where each column shows one
weighting map $\tilde{A}_i$ for generating one holistic codeword. Some weighting maps focus on summarizing  foreground objects or regions to create holistic codewords, while some other weighting maps pay attention to summarizing background contextual regions or objects as the holistic codewords. The visualization shows that the learned codewords implicitly capture different global contexts from the scenes.
In Fig.~\ref{fig:vis}, we also visualize some predictions by the baseline
DilatedFCN-8s and by our EfficientFCN, where our model significantly improves the visualized results with the proposed HGD.

\begin{table*}[!t]
	\centering
	\caption{The effects of different dimensions of holistic codewords in our HGD-FPN when integrated in Faster RCNN with a ResNet-50 backbone on MS COCO 2017 \emph{minival}. The number of recurrence of HGD $k=4$.}
	\addtolength{\tabcolsep}{-1pt}
	\small{
		\begin{tabular}{c|c|c|cc|ccccccc}
			\hline
			\multirow{2}{*}{Feature Fusion} &\multirow{2}{*}{$n$} & \multirow{2}{*}{$c$}&\multirow{2}{*}{FLOPs (G)} &\multirow{2}{*}{Params (M)} &  \multirow{2}{*}{Task} & \multirow{2}{*}{AP} & \multirow{2}{*}{$\text{AP}_{50}$} & \multirow{2}{*}{$\text{AP}_{75}$} & \multirow{2}{*}{$\text{AP}_{S}$} & \multirow{2}{*}{$\text{AP}_{M}$} & \multirow{2}{*}{$\text{AP}_{L}$} \\
			&  &  &  &  &  &  &  &  \\ \hline
			 FPN \cite{lin2017feature} & - & - & 214.8 & 41.5 & BBox & 36.3 & 58.2 & 39.1 & 21.6 & 40.1 & 46.4 \\ \hline
			 \multirow{10}{*}{HGD-FPN} & 0 & 0 &  266.4 & 43.3 &BBox  & {37.5} & {59.2} & {40.3} & {21.6} & {41.0} & {48.6} \\ \cline{2-12}
			  & \multirow{3}{*} {64} & 256 & 384.9 & 47.9 & BBox  & {39.4} & {61.3} & {42.4} & {24.6} & {42.5} & {51.0} \\ 
		          &  & 512 &  554.8  & 54.2 & BBox &{39.7} & {61.6} & {42.7} & {24.4} & {43.3} & {50.9} \\ \
		          &  & 1024  & 894.8 & 66.9 & BBox & {40.1} & {61.8} & {43.7} & {24.5} & {43.9} & {51.0} \\ \cline{2-12}
		          & \multirow{3}{*}{128} & 256 & 397.8 & 48.4 & BBox & {39.4} & {61.1} & {42.6} & {24.0} & {42.9} & {50.8} \\ 
		          &  & 512 & 580.7 & 55.1 & BBox & {40.0} & {61.9} & {43.6} & {24.5} & {43.6} & {51.4} \\ 
		          &  & 1024 & 946.4 & 68.7 & BBox & {39.7} & {61.5} & {42.9} & {24.1} & {43.1} & {51.3} \\ \cline{2-12}
		          & \multirow{3}{*}{256} & 256 & 423.8 & 49.7 & BBox & {39.5} & {61.2} & {42.7} & {23.9} & {43.0} & {50.7} \\ 
		          &  & 512 & 632.4 & 56.9 & BBox & {39.7} & {61.6} & {43.0} & {24.3} & {43.2} & {51.4} \\ 
		          &  & 1024 & 1049.6 & 72.3 & BBox & {40.0} & {61.9} & {43.3} & {24.7} & {43.6} & {51.9} \\ \hline
		\end{tabular}
	}
	\label{tab:ablation_nc}
\end{table*}

\subsubsection{Results on PASCAL VOC}
The original PASCAL VOC 2012 dataset consists of 1,464 images for training, 1,449 for validation, and 1,456 for testing, which is a major benchmark dataset for semantic object segmentation. It includes 20 foreground objects classed and one background class. The augmented training set of 10,582 images, namely train-aug, is adopted as the training set following the previous experimental set in \cite{Zhang_2019_CVPR}.
To further demonstrate the effectiveness of our proposed HGD. We adopt all the best strategies of HGD design and compare it with state-of-the-art methods on the test set of PASCAL-VOC 2012, which is evaluated on the official online server. As shown in Table \ref{table:pascal_voc_2012}, the dilatedFCN based methods dominate the top performances on the PASCAL VOC benchmark. However, our EfficientFCN with a backbone having no dilated convolution can still achieve the best results among all the ResNet101-based methods. 
\subsubsection{Results on ADE20K}
The ADE20K dataset consists of 20K images for training, 2K images for
validation, and 3K images for testing, which were used for ImageNet Scene
Parsing Challenge 2016. This dataset is more complex and challenging with 150 labeled classes and more diverse scenes. As shown in Table \ref{table:pascal-context}, our
EfficientFCN achieves the competitive performance than the dilatedFCN based
methods but has only 1/3 of their computational cost.

\begin{table*}[!t]
	\centering
	\caption{The effects of integrating different FPN variants in Faster RCNN with a ResNet-50 backbone on MS COCO 2017 \emph{minival}.}
	\addtolength{\tabcolsep}{-1pt}
	\small{
		\begin{tabular}{c|l|cc|lllllll}
			\hline
			\multirow{2}{*}{Feature Fusion} &\multirow{2}{*}{$k$} &\multirow{2}{*}{FLOPs (G)} &\multirow{2}{*}{Params (M)} & \multirow{2}{*}{Task} & \multirow{2}{*}{AP} & \multirow{2}{*}{$\text{AP}_{50}$} & \multirow{2}{*}{$\text{AP}_{75}$} & \multirow{2}{*}{$\text{AP}_{S}$} & \multirow{2}{*}{$\text{AP}_{M}$} & \multirow{2}{*}{$\text{AP}_{L}$} \\
			&   &  &  &  &  &  &  \\ \hline
			 FPN \cite{lin2017feature}  & 1  & 214.8 & 41.5 & BBox & 36.3 & 58.2 & 39.1 & 21.6 & 40.1 & 46.4 \\ \hline
			 CARAFE \cite{wang2019carafe} & 1  & 217.8 & 47.1 & BBox & {38.1} & {60.7} & {41.0} & {22.8} & {41.2} & {46.9} \\ \hline
			 NAS-FPN \cite{ghiasi2019fpn} & 7  & 666.9 & 68.2 & BBox & {39.0} & {59.5} & {42.4} & {22.4} & {42.6} & {47.8} \\ \hline
			 FPG \cite{chen2020feature} & 9 & 637.8 & 79.3 & BBox & {39.2} & {60.8} & {42.7} & {22.7} & {41.9} & {48.4} \\ \hline
			 \multirow{5}{*}{HGD-FPN} & 1 & 306.3 & 55.1 & BBox  & {37.9} & {60.7} & {40.6} & {23.8} & {42.0} & {48.1} \\
		          & 2  & 397.7 & 55.1 & BBox & {38.6} & {60.9} & {41.9} & {24.0} & {42.4} & {49.4} \\
		          & 3  & 489.2 & 55.1 & BBox & {39.3} & {61.5} & {42.6} & {24.2} & {42.8} & {50.6} \\ 
		          & 4  & 580.7 & 55.1 &BBox & {40.0} & {61.9} & {43.6} & {24.5} & {43.6} & {51.4} \\
		          & 5  & 672.1 & 55.1 &BBox & {39.8} & {61.4} & {42.9} & {23.4} & {43.4} & {51.9} \\ \hline
		\end{tabular}
	}
	\label{tab:ablation_k}
\end{table*}

\subsection{Object Detection and Instance Segmentation}

We verify the effectiveness of our proposed HGD-FPN to generate multi-scale feature pyramids for object detection and instance segmentation on COCO 2017 detection dataset \cite{lin2014microsoft}.
This dataset contains objects of 80 categories. The training dataset consists of $\sim$118k samples, and the validation dataset contains 5k images (\emph{minival}).
The performances on the \emph{minival} set are reported. The evaluation metrics include Average Precision (AP), AP$_{50}$ (AP for IoU threshold 50$\%$),
AP$_{75}$ (AP for IoU threshold 75$\%$), AP$_{S}$ (results on small scale), AP$_{M}$ (results on medium scale) and AP$_{L}$ (results on large scale).

We integrate and test our HGD-FPN in three object detection frameworks (Faster RCNN \cite{ren2015faster}, RetinaNet \cite{lin2017focal} and FreeAnchor \cite{zhang2019freeanchor}), and 
one framework for instance segmentation (Mask RCNN \cite{he2017mask}) with ResNet-50 \cite{he2016deep} backbones. The 1x training schedule settings are adopted following Detectron \cite{Detectron2018} and MMDetection \cite{chen2019mmdetection}. For fair comparison, all experiments are conducted based on MMDetection \cite{chen2019mmdetection} implementations.
The default parameters for our HGD-FPN are set as: the number of codewords $n=128$, the dimension of codewords $c=512$, and the FPN recurrence number $k=4$. 
The HGD-FPN module shares its parameters at different recurrence stages.

\subsubsection{Ablation Studies}

To investigate the effects of different components of our proposed HGD-FPN, we design a series of experiments on HGD-FPN integrated in the Faster RCNN \cite{ren2015faster} framework.

\noindent \textbf{Dimensions of the shared holistic codewords.}
For our HGD-FPN, the core contribution is to generate the multi-scale feature maps via a linear combination of the 
the predicted holistic codewords. Then, the number of codewords $n$ and the dimension of the codewords $c$ have major impacts to the performance of the proposed HGD-FPN. As the channel of the input feature pyramids $\{P3-P7\}$ have been reduced to 256 \cite{ren2015faster, lin2017focal}, 
we design a series of experiments with varying $n$ and $c$ to investigate their influences in Table. \ref{tab:ablation_nc}.  

The implementation with $n=0$ and $c=0$ means keeping only the multi-feature fusion operation while removing the holistic codeword generation and codeword assembly operations in our HGD module. 
The HGD-FPN with $n=0$ and $c=0$ would still lead to a better result than the original FPN method.
Second, compared with the performance of our HGD-FPN with $n=0$ and $c=0$, the results of our HGD-FPN with $n \geq 1$ and $c \geq 1$ show significant improvements, 
which demonstrate that our proposed HGD module can generate discriminative multi-scale feature maps for object detection. 
Finally, as $n$ and $c$ increase, the better performance can be achieved with higher computational cost.
We choose $n=128$ and $c=512$ as our default parameter settings in the following experiments as trade-offs between the performance and the computational efficiency. 

\noindent \textbf{The recurrence number $k$ of HGD-FPN.}
With the fixed dimensions for the holistic codewords, we further investigate the effect of the recurrence number $k$ of the proposed HGD-FPN module.
We conduct five experiments with varying $k$'s from 1 to 5. As shown in the Table. \ref{tab:ablation_k}, increasing the recurrence number $k$ of HGD-FPN results in better performance, which demonstrates that the recurrence of the HGD-FPN module could further strengthen the multi-scale feature fusion and increase the discriminativeness of the feature pyramids. Even when $k=1$, the performance of our HGD-FPN is still better than that of conventional FPN \cite{lin2017feature} and similar to that of CARAFE \cite{wang2019carafe}.
Although the method CARAFE\cite{wang2019carafe} has a very smaller computational cost than our HGD-FPN with $k=1$, the method CARAFE is only designed for context-aware feature upsampling operation which can be integrated into the FPN module in the top-down path and  cannot be used for the feature fusion from the bottom-up path. Then, similar to the weakness of the classical FPN method \cite{ghiasi2019fpn}, we found that stacking the CARAFE architecture does not improve the performance.
The FPN variants, NAS-FPN \cite{ghiasi2019fpn} and FPG \cite{chen2020feature}, both investigate recurrent or multiple repeated FPN architectures. NAS-FPN \cite{ghiasi2019fpn} stacks the searched FPN architecture and FPG \cite{chen2020feature} stacks the fully-connected FPN architecture. In Table. \ref{tab:ablation_k}, we observe that our proposed HGD-FPN ($k=4$) has higher accuracy than NAS-FPN ($k=7$) \cite{ghiasi2019fpn} and FPG ($k=9$) \cite{chen2020feature} with less computational cost.
In the remaining parts of the paper, we set the default $k$ as 4.

\noindent \textbf{Parameter sharing of the recurrent HGD-FPN module.}
We also investigate the effect of whether to share parameters of the recurrent HGD-FPN at different stages. For the shared-parameter setting, we only initialize one set of parameters for our HGD-FPN and stack it for $k$ times. For the non-sharing parameter setting, we use $k$ different sets of parameters for HGD-FPN. In such cases, for $k=4$, the mAP of the Faster RCNN with a ResNet 50 backbone is 39.6, which is smaller than that of the method with shared parameters. Then, we choose the parameter sharing strategy for our proposed HGD-FPN to reduce the memory cost.

\begin{table*}[!t]
	\centering
	\caption{{Object Detection performances of different FPN variants and frameworks on COCO 2017 \emph{minival} set.}}
	\addtolength{\tabcolsep}{-2pt}
	\small{
		\begin{tabular}{l|l|llllllll}
			\hline
			\multirow{2}{*}{Method} & \multirow{2}{*}{Feature Fusion} & \multirow{2}{*}{Backbone} & \multirow{2}{*}{Task} & \multirow{2}{*}{AP} & \multirow{2}{*}{$\text{AP}_{50}$} & \multirow{2}{*}{$\text{AP}_{75}$} & \multirow{2}{*}{$\text{AP}_{S}$} & \multirow{2}{*}{$\text{AP}_{M}$} & \multirow{2}{*}{$\text{AP}_{L}$} \\
			&  &  &  &  &  &  &  &  \\ \hline
			\multirow{6}{*}{Faster R-CNN \cite{ren2015faster}} & FPN \cite{lin2017feature} & ResNet-50 & BBox & 36.3 & 58.2 & 39.1 & 21.6 & 40.1 & 46.4 \\
			 & PA-FPN \cite{liu2018path} & ResNet-50 & BBox & {36.7} & {58.5} & {39.7} & {21.6} & {40.4} & {47.7} \\
		         & Balanced FPN \cite{pang2019libra} & ResNet-50 & BBox & {37.2} & {59.8} & {40.1} & {23.2} & {41.2} & {47.1} \\\
			 & CARAFE \cite{wang2019carafe} & ResNet-50 & BBox & {38.1} & {60.7} & {41.0} & {22.8} & {41.2} & {46.9} \\
			 & HGD-FPN & ResNet-50 & BBox & \textbf{40.0} & \textbf{61.9} & \textbf{43.6} & \textbf{24.5} & \textbf{43.6} & \textbf{51.4} \\ \hline
			\multirow{5}{*}{RetinaNet \cite{lin2017focal}} & FPN \cite{lin2017feature} & ResNet-50 & BBox & 35.3 & 55.1 & 37.7 & 19.6 & 39.2 & 46.9 \\
			 & PA-FPN \cite{liu2018path} & ResNet-50 & BBox & {35.8} & {55.6} & {38.3} & {20.1} & {39.9} & {47.1} \\
		         & Balanced FPN \cite{pang2019libra} & ResNet-50 & BBox & {36.4} & {57.2} & {38.9} & {21.6} & {40.5} & {47.0} \\
			 & CARAFE \cite{wang2019carafe} & ResNet-50 & BBox & {36.1} & {56.5} & {38.3} & {20.9} & {40.3} & {46.8} \\
			 & HGD-FPN & ResNet-50 & BBox & \textbf{37.8} & \textbf{58.2} & \textbf{40.7} & \textbf{22.0} & \textbf{42.0} & \textbf{49.6} \\ \hline
			\multirow{5}{*}{FreeAnchor \cite{zhang2019freeanchor}} & FPN \cite{lin2017feature} & ResNet-50 & BBox & 38.5 & 57.3 & 41.2 & 21.1 & 41.8 & 51.5 \\
			 & PA-FPN \cite{liu2018path} & ResNet-50 & BBox & {38.9} & {57.6} & {41.6} & {21.8} & {41.8} & {50.9} \\
		         & Balanced FPN \cite{pang2019libra} & ResNet-50 & BBox & {39.1} & {58.5} & {41.6} & {22.0} & {42.9} & {51.4} \\
			 & CARAFE \cite{wang2019carafe} & ResNet-50 & BBox & {39.2} & {58.4} & {42.0} & {22.5} & {42.9} & {51.5} \\
			 & HGD-FPN & ResNet-50 & BBox & \textbf{40.8} & \textbf{59.7} & \textbf{43.7} & \textbf{23.2} & \textbf{44.0} & \textbf{54.5} \\ \hline
		\end{tabular}
	}
	\label{tab:det-results}
\end{table*}

\begin{table*}[!t]
	\centering
	\caption{{Instance Segmentation results on COCO 2017 \emph{minival} set.}}
	\addtolength{\tabcolsep}{-2pt}
	\small{
		\begin{tabular}{l|l|llllllll}
			\hline
			\multirow{2}{*}{Method} & \multirow{2}{*}{Feature Fusion} & \multirow{2}{*}{Backbone} & \multirow{2}{*}{Task} & \multirow{2}{*}{AP} & \multirow{2}{*}{$\text{AP}_{50}$} & \multirow{2}{*}{$\text{AP}_{75}$} & \multirow{2}{*}{$\text{AP}_{S}$} & \multirow{2}{*}{$\text{AP}_{M}$} & \multirow{2}{*}{$\text{AP}_{L}$} \\
			&  &  &  &  &  &  &  &  \\ \hline
			\multirow{10}{*}{Mask R-CNN \cite{he2017mask}} & \multirow{2}{*}{FPN \cite{lin2017feature}} & ResNet-50 & BBox & 37.0 & 58.9 & 40.0 & 22.3 & 40.6 & 47.7 \\
			& & ResNet-50 & Segm & 34.1 & 55.6 & 35.9 & 18.8 & 37.2 & 46.3 \\
			\cline{2-10}
                        & \multirow{2}{*}{PA-FPN \cite{liu2018path}} & ResNet-50 & BBox & {37.5} & {59.2} & {40.6} & {22.9} & {41.3} & {48.5} \\
			& & ResNet-50 & Segm & {34.5}  & {55.9} & {36.6} & {18.4} & {37.9} & {47.2} \\
			\cline{2-10}
                        & \multirow{2}{*}{Balanced FPN \cite{pang2019libra}} & ResNet-50 & BBox & {38.2} & {60.6} & {41.2} & {23.2} & {42.1} & {49.3} \\
			& & ResNet-50 & Segm & {34.9} & {57.0} &  {37.1} &{19.3} & {38.5} & {47.6} \\
			\cline{2-10}
                        & \multirow{2}{*}{CARAFE \cite{wang2019carafe}} & ResNet-50 & BBox & {38.8} & {61.2} & {42.1} & {23.2} & {41.7} & {47.9} \\
			& & ResNet-50 & Segm & {35.9} & {58.1} & {38.2} & {19.8} & {38.6} & {46.5} \\
			\cline{2-10}
			& \multirow{2}{*}{HGD-FPN} & ResNet-50 & BBox & \textbf{40.8} & \textbf{61.9} & \textbf{44.6} & \textbf{25.1} & \textbf{44.4} & \textbf{52.0} \\
			& & ResNet-50 & Segm & \textbf{36.5} & \textbf{58.5} & \textbf{38.9} & \textbf{20.5} & \textbf{39.9} & \textbf{49.2} \\ \hline
		\end{tabular}
	}
	\label{tab:instance-results}
\end{table*}

\subsubsection{Comparison with state-of-the-art FPN variants}
\label{ssec:comparison_fpn}

To further demonstrate the effectiveness of our proposed HGD-FPN for object detection, we integrate it into two other classical object detection frameworks, RetinaNet \cite{lin2017focal} and FreeAnchor \cite{zhang2019freeanchor}. Different from Faster RCNN \cite{ren2015faster}, which is a two-stage object detector, RetinaNet \cite{lin2017focal} and FreeAnchor \cite{zhang2019freeanchor} are one-stage object detection frameworks. 
Our proposed HGD-FPN is compared with state-of-the-art multi-scale FPN variants within such object detection frameworks.

As shown in Table \ref{tab:det-results}, our proposed HGD-FPN achieves the best performance when integrated into the three object detection frameworks.
Both the conventional FPN \cite{lin2017feature} and PA-FPN \cite{liu2018path} obtain the upsampled feature maps from local neighborhoods, which lack the ability of capturing long-range context information. 
Although CARAFE \cite{wang2019carafe} uses an adaptive feature upsampling scheme, the receptive fields of its predicted kernels are still limited.
For Balanced FPN \cite{pang2019libra}, it adopts the non-local module to generate balanced multi-scale features for enhancing the output feature pyramids. However, such features might not be able to capture the global context for each feature scale.
Since the holistic codewords in HGD-FPN are generated by considering the holistic information and the multi-scale feature maps are generated via linearly combining them with predicted assembly coefficients, our HGD-FPN could adaptively enhance each feature scale with the global contexts and improve the final performance for object detection. 

Besides object detection, we integrate our proposed HGD-FPN into MaskRCNN for instance segmentation. As shown in Table \ref{tab:instance-results}, our HGD-FPN also leads to the most AP gain compared to other FPN variants when integrated with Mask RCNN \cite{he2017mask}. The good performance validates that the proposed HGD could yield the discriminative semantic feature maps and help boost the performance of instance segmentation.

\section{Conclusions}
In this paper, we propose the holistically-guided decoder for achieving discriminative deep feature representations. In fact, this novel decoder is flexible to reconstruct semantic-rich feature maps of any resolution from the holistic information in the feature pyramids of the encoder. To validate the effectiveness of its feature upsampling performance, we test it in two computer vision tasks and design two instantiations for semantic segmentation and the object detection/instance segmentation. With the proposed HGD, our EfficientFCN, with much fewer parameters and less computational cost, achieves competitive or even better performance compared with state-of-the-art dilatedFCN based methods. For object detection/instance segmentation, we propose a novel HGD-FPN to produce multi-scale feature pyramids. Both the proposed EfficientFCN and the HGD-FPN architectures demonstrate that our proposed HGD has powerful capability of encoding discriminative multi-scale feature representations.

{\small
\bibliographystyle{IEEEtran}
\bibliography{hgd_feature_learning}

\begin{thebibliography}{10}
\providecommand{\url}[1]{#1}
\csname url@samestyle\endcsname
\providecommand{\newblock}{\relax}
\providecommand{\bibinfo}[2]{#2}
\providecommand{\BIBentrySTDinterwordspacing}{\spaceskip=0pt\relax}
\providecommand{\BIBentryALTinterwordstretchfactor}{4}
\providecommand{\BIBentryALTinterwordspacing}{\spaceskip=\fontdimen2\font plus
\BIBentryALTinterwordstretchfactor\fontdimen3\font minus
  \fontdimen4\font\relax}
\providecommand{\BIBforeignlanguage}[2]{{%
\expandafter\ifx\csname l@#1\endcsname\relax
\typeout{** WARNING: IEEEtran.bst: No hyphenation pattern has been}%
\typeout{** loaded for the language `#1'. Using the pattern for}%
\typeout{** the default language instead.}%
\else
\language=\csname l@#1\endcsname
\fi
#2}}
\providecommand{\BIBdecl}{\relax}
\BIBdecl

\bibitem{chen2017deeplab}
L.-C. Chen, G.~Papandreou, I.~Kokkinos, K.~Murphy, and A.~L. Yuille, ``Deeplab:
  Semantic image segmentation with deep convolutional nets, atrous convolution,
  and fully connected crfs,'' \emph{IEEE transactions on pattern analysis and
  machine intelligence}, vol.~40, no.~4, pp. 834--848, 2017.

\bibitem{chen2017rethinking}
L.-C. Chen, G.~Papandreou, F.~Schroff, and H.~Adam, ``Rethinking atrous
  convolution for semantic image segmentation,'' \emph{arXiv preprint
  arXiv:1706.05587}, 2017.

\bibitem{yu2017dilated}
F.~Yu, V.~Koltun, and T.~Funkhouser, ``Dilated residual networks,'' in
  \emph{Proceedings of the IEEE conference on computer vision and pattern
  recognition}, 2017, pp. 472--480.

\bibitem{ronneberger2015u}
O.~Ronneberger, P.~Fischer, and T.~Brox, ``U-net: Convolutional networks for
  biomedical image segmentation,'' in \emph{International Conference on Medical
  image computing and computer-assisted intervention}.\hskip 1em plus 0.5em
  minus 0.4em\relax Springer, 2015, pp. 234--241.

\bibitem{Zhang_2018_CVPR}
H.~Zhang, K.~Dana, J.~Shi, Z.~Zhang, X.~Wang, A.~Tyagi, and A.~Agrawal,
  ``Context encoding for semantic segmentation,'' in \emph{The IEEE Conference
  on Computer Vision and Pattern Recognition (CVPR)}, June 2018.

\bibitem{he2019adaptive}
J.~He, Z.~Deng, L.~Zhou, Y.~Wang, and Y.~Qiao, ``Adaptive pyramid context
  network for semantic segmentation,'' in \emph{Proceedings of the IEEE
  Conference on Computer Vision and Pattern Recognition}, 2019, pp. 7519--7528.

\bibitem{Zhang_2019_CVPR}
H.~Zhang, H.~Zhang, C.~Wang, and J.~Xie, ``Co-occurrent features in semantic
  segmentation,'' in \emph{The IEEE Conference on Computer Vision and Pattern
  Recognition (CVPR)}, 2019.

\bibitem{SegNet}
V.~Badrinarayanan, A.~Kendall, and R.~Cipolla, ``Segnet: A deep convolutional
  encoder-decoder architecture for image segmentation,'' \emph{arXiv preprint
  arXiv:1511.00561}, 2015.

\bibitem{lin2017feature}
T.-Y. Lin, P.~Doll{\'a}r, R.~Girshick, K.~He, B.~Hariharan, and S.~Belongie,
  ``Feature pyramid networks for object detection,'' in \emph{Proceedings of
  the IEEE conference on computer vision and pattern recognition}, 2017, pp.
  2117--2125.

\bibitem{liu2018path}
S.~Liu, L.~Qi, H.~Qin, J.~Shi, and J.~Jia, ``Path aggregation network for
  instance segmentation,'' in \emph{Proceedings of the IEEE conference on
  computer vision and pattern recognition}, 2018, pp. 8759--8768.

\bibitem{ghiasi2019fpn}
G.~Ghiasi, T.-Y. Lin, and Q.~V. Le, ``Nas-fpn: Learning scalable feature
  pyramid architecture for object detection,'' in \emph{Proceedings of the IEEE
  conference on computer vision and pattern recognition}, 2019, pp. 7036--7045.

\bibitem{ren2015faster}
S.~Ren, K.~He, R.~Girshick, and J.~Sun, ``Faster r-cnn: Towards real-time
  object detection with region proposal networks,'' in \emph{Advances in neural
  information processing systems}, 2015, pp. 91--99.

\bibitem{lin2017focal}
T.-Y. Lin, P.~Goyal, R.~Girshick, K.~He, and P.~Doll{\'a}r, ``Focal loss for
  dense object detection,'' in \emph{Proceedings of the IEEE international
  conference on computer vision}, 2017, pp. 2980--2988.

\bibitem{zhang2019freeanchor}
X.~Zhang, F.~Wan, C.~Liu, R.~Ji, and Q.~Ye, ``Freeanchor: Learning to match
  anchors for visual object detection,'' in \emph{Advances in Neural
  Information Processing Systems}, 2019, pp. 147--155.

\bibitem{he2017mask}
K.~He, G.~Gkioxari, P.~Doll{\'a}r, and R.~Girshick, ``Mask r-cnn,'' in
  \emph{Proceedings of the IEEE international conference on computer vision},
  2017, pp. 2961--2969.

\bibitem{fu2019dual}
J.~Fu, J.~Liu, H.~Tian, Y.~Li, Y.~Bao, Z.~Fang, and H.~Lu, ``Dual attention
  network for scene segmentation,'' in \emph{Proceedings of the IEEE Conference
  on Computer Vision and Pattern Recognition}, 2019, pp. 3146--3154.

\bibitem{Fu_2019_ICCV}
J.~Fu, J.~Liu, Y.~Wang, Y.~Li, Y.~Bao, J.~Tang, and H.~Lu, ``Adaptive context
  network for scene parsing,'' in \emph{The IEEE International Conference on
  Computer Vision (ICCV)}, October 2019.

\bibitem{he2019dynamic}
J.~He, Z.~Deng, and Y.~Qiao, ``Dynamic multi-scale filters for semantic
  segmentation,'' in \emph{Proceedings of the IEEE International Conference on
  Computer Vision}, 2019, pp. 3562--3572.

\bibitem{zhao2017pyramid}
H.~Zhao, J.~Shi, X.~Qi, X.~Wang, and J.~Jia, ``Pyramid scene parsing network,''
  in \emph{Proceedings of the IEEE conference on computer vision and pattern
  recognition}, 2017, pp. 2881--2890.

\bibitem{takikawa2019gated}
T.~Takikawa, D.~Acuna, V.~Jampani, and S.~Fidler, ``Gated-scnn: Gated shape
  cnns for semantic segmentation,'' in \emph{Proceedings of the IEEE
  International Conference on Computer Vision}, 2019, pp. 5229--5238.

\bibitem{tian2019decoders}
Z.~Tian, T.~He, C.~Shen, and Y.~Yan, ``Decoders matter for semantic
  segmentation: Data-dependent decoding enables flexible feature aggregation,''
  in \emph{Proceedings of the IEEE Conference on Computer Vision and Pattern
  Recognition}, 2019, pp. 3126--3135.

\bibitem{wu2019fastfcn}
H.~Wu, J.~Zhang, K.~Huang, K.~Liang, and Y.~Yu, ``Fastfcn: Rethinking dilated
  convolution in the backbone for semantic segmentation,'' \emph{arXiv preprint
  arXiv:1903.11816}, 2019.

\bibitem{girshick2015fast}
R.~Girshick, ``Fast r-cnn,'' in \emph{Proceedings of the IEEE international
  conference on computer vision}, 2015, pp. 1440--1448.

\bibitem{singh2018analysis}
B.~Singh and L.~S. Davis, ``An analysis of scale invariance in object detection
  snip,'' in \emph{Proceedings of the IEEE conference on computer vision and
  pattern recognition}, 2018, pp. 3578--3587.

\bibitem{singh2018sniper}
B.~Singh, M.~Najibi, and L.~S. Davis, ``Sniper: Efficient multi-scale
  training,'' in \emph{Advances in neural information processing systems},
  2018, pp. 9310--9320.

\bibitem{pang2019libra}
J.~Pang, K.~Chen, J.~Shi, H.~Feng, W.~Ouyang, and D.~Lin, ``Libra r-cnn:
  Towards balanced learning for object detection,'' in \emph{Proceedings of the
  IEEE conference on computer vision and pattern recognition}, 2019, pp.
  821--830.

\bibitem{wang2019carafe}
J.~Wang, K.~Chen, R.~Xu, Z.~Liu, C.~C. Loy, and D.~Lin, ``Carafe: Content-aware
  reassembly of features,'' in \emph{Proceedings of the IEEE International
  Conference on Computer Vision}, 2019, pp. 3007--3016.

\bibitem{chen2020feature}
K.~Chen, Y.~Cao, C.~C. Loy, D.~Lin, and C.~Feichtenhofer, ``Feature pyramid
  grids,'' \emph{arXiv preprint arXiv:2004.03580}, 2020.

\bibitem{hu2018squeeze}
J.~Hu, L.~Shen, and G.~Sun, ``Squeeze-and-excitation networks,'' in
  \emph{Proceedings of the IEEE conference on computer vision and pattern
  recognition}, 2018, pp. 7132--7141.

\bibitem{he2016deep}
K.~He, X.~Zhang, S.~Ren, and J.~Sun, ``Deep residual learning for image
  recognition,'' in \emph{Proceedings of the IEEE conference on computer vision
  and pattern recognition}, 2016, pp. 770--778.

\bibitem{tan2020efficientdet}
M.~Tan, R.~Pang, and Q.~V. Le, ``Efficientdet: Scalable and efficient object
  detection,'' in \emph{Proceedings of the IEEE/CVF Conference on Computer
  Vision and Pattern Recognition}, 2020, pp. 10\,781--10\,790.

\bibitem{mottaghi2014role}
R.~Mottaghi, X.~Chen, X.~Liu, N.-G. Cho, S.-W. Lee, S.~Fidler, R.~Urtasun, and
  A.~Yuille, ``The role of context for object detection and semantic
  segmentation in the wild,'' in \emph{Proceedings of the IEEE Conference on
  Computer Vision and Pattern Recognition}, 2014, pp. 891--898.

\bibitem{everingham2010pascal}
M.~Everingham, L.~Van~Gool, C.~K. Williams, J.~Winn, and A.~Zisserman, ``The
  pascal visual object classes (voc) challenge,'' \emph{International journal
  of computer vision}, vol.~88, no.~2, pp. 303--338, 2010.

\bibitem{zhou2017scene}
B.~Zhou, H.~Zhao, X.~Puig, S.~Fidler, A.~Barriuso, and A.~Torralba, ``Scene
  parsing through ade20k dataset,'' in \emph{Proceedings of the IEEE conference
  on computer vision and pattern recognition}, 2017, pp. 633--641.

\bibitem{lin2014microsoft}
T.-Y. Lin, M.~Maire, S.~Belongie, J.~Hays, P.~Perona, D.~Ramanan,
  P.~Doll{\'a}r, and C.~L. Zitnick, ``Microsoft coco: Common objects in
  context,'' in \emph{European conference on computer vision}.\hskip 1em plus
  0.5em minus 0.4em\relax Springer, 2014, pp. 740--755.

\bibitem{russakovsky2015imagenet}
O.~Russakovsky, J.~Deng, H.~Su, J.~Krause, S.~Satheesh, S.~Ma, Z.~Huang,
  A.~Karpathy, A.~Khosla, M.~Bernstein \emph{et~al.}, ``Imagenet large scale
  visual recognition challenge,'' \emph{International journal of computer
  vision}, vol. 115, no.~3, pp. 211--252, 2015.

\bibitem{bottou2010large}
L.~Bottou, ``Large-scale machine learning with stochastic gradient descent,''
  in \emph{Proceedings of COMPSTAT'2010}.\hskip 1em plus 0.5em minus
  0.4em\relax Springer, 2010, pp. 177--186.

\bibitem{RefineNet}
G.~Lin, A.~Milan, C.~Shen, and I.~Reid, ``Refinenet: Multi-path refinement
  networks for high-resolution semantic segmentation,'' in \emph{Proceedings of
  the IEEE conference on computer vision and pattern recognition}, 2017, pp.
  1925--1934.

\bibitem{MSCI}
D.~Lin, Y.~Ji, D.~Lischinski, D.~Cohen-Or, and H.~Huang, ``Multi-scale context
  intertwining for semantic segmentation,'' in \emph{Proceedings of the
  European Conference on Computer Vision (ECCV)}, 2018, pp. 603--619.

\bibitem{zhang2017scale}
R.~Zhang, S.~Tang, Y.~Zhang, J.~Li, and S.~Yan, ``Scale-adaptive convolutions
  for scene parsing,'' in \emph{Proceedings of the IEEE International
  Conference on Computer Vision}, 2017, pp. 2031--2039.

\bibitem{zhu2019asymmetric}
Z.~Zhu, M.~Xu, S.~Bai, T.~Huang, and X.~Bai, ``Asymmetric non-local neural
  networks for semantic segmentation,'' in \emph{Proceedings of the IEEE
  International Conference on Computer Vision}, 2019, pp. 593--602.

\bibitem{long2015fully}
J.~Long, E.~Shelhamer, and T.~Darrell, ``Fully convolutional networks for
  semantic segmentation,'' in \emph{Proceedings of the IEEE conference on
  computer vision and pattern recognition}, 2015, pp. 3431--3440.

\bibitem{CRF-RNN}
S.~Zheng, S.~Jayasumana, B.~Romera-Paredes, V.~Vineet, Z.~Su, D.~Du, C.~Huang,
  and P.~H. Torr, ``Conditional random fields as recurrent neural networks,''
  in \emph{Proceedings of the IEEE international conference on computer
  vision}, 2015, pp. 1529--1537.

\bibitem{DeconvNet}
H.~Noh, S.~Hong, and B.~Han, ``Learning deconvolution network for semantic
  segmentation,'' in \emph{Proceedings of the IEEE international conference on
  computer vision}, 2015, pp. 1520--1528.

\bibitem{DPN}
Z.~Liu, X.~Li, P.~Luo, C.-C. Loy, and X.~Tang, ``Semantic image segmentation
  via deep parsing network,'' in \emph{Proceedings of the IEEE International
  Conference on Computer Vision}, 2015, pp. 1377--1385.

\bibitem{Piecewise}
G.~Lin, C.~Shen, A.~Van Den~Hengel, and I.~Reid, ``Efficient piecewise training
  of deep structured models for semantic segmentation,'' in \emph{Proceedings
  of the IEEE Conference on Computer Vision and Pattern Recognition}, 2016, pp.
  3194--3203.

\bibitem{ResNet38}
Z.~Wu, C.~Shen, and A.~v.~d. Hengel, ``Wider or deeper: Revisiting the resnet
  model for visual recognition,'' \emph{arXiv preprint arXiv:1611.10080}, 2016.

\bibitem{Detectron2018}
R.~Girshick, I.~Radosavovic, G.~Gkioxari, P.~Doll\'{a}r, and K.~He,
  ``Detectron,'' \url{https://github.com/facebookresearch/detectron}, 2018.

\bibitem{chen2019mmdetection}
K.~Chen, J.~Wang, J.~Pang, Y.~Cao, Y.~Xiong, X.~Li, S.~Sun, W.~Feng, Z.~Liu,
  J.~Xu \emph{et~al.}, ``Mmdetection: Open mmlab detection toolbox and
  benchmark,'' \emph{arXiv preprint arXiv:1906.07155}, 2019.

\end{thebibliography}
}

\end{document}